\documentclass[10pt,twocolumn,letterpaper]{article}

\usepackage{cvpr}
\usepackage[pagebackref=true,breaklinks=true,letterpaper=true,colorlinks,bookmarks=false]{hyperref}
\usepackage{times}
\usepackage{epsfig}
\usepackage[margin=20pt]{subfig}
\usepackage{graphicx,amsmath,amssymb,booktabs,tabulary,multirow,overpic,xcolor}
\usepackage[british,english]{babel}
\usepackage{soul}
\usepackage{xspace}\xspace
\usepackage{adjustbox}
\usepackage{array}
\usepackage{bm}

\usepackage{dblfloatfix}
\usepackage{transparent}

\usepackage{algorithm}
\usepackage{listings}

\usepackage{microtype}

\newcommand*\mat[1]{\mathbf{#1}}
\newcommand*\vv[1]{\mathbf{#1}}

\newcommand*\trp[1]{ #1^{\mathsf{T}} }

\newcommand{\vx}{\vv{x}}

\DeclareMathOperator*{\concat}{||}

\newcommand*\mypar[1]{\vspace{0.5em} \noindent \textbf{#1} \hspace{0.5em}}

\newcommand{\bd}[1]{\textbf{#1}}
\newcommand{\app}{\raise.17ex\hbox{$\scriptstyle\sim$}}

\newcolumntype{x}[1]{>{\centering\arraybackslash}p{#1pt}}

\newlength\savewidth\newcommand\shline{\noalign{\global\savewidth\arrayrulewidth
  \global\arrayrulewidth 1pt}\hline\noalign{\global\arrayrulewidth\savewidth}}
\newcommand{\tablestyle}[2]{\setlength{\tabcolsep}{#1}\renewcommand{\arraystretch}{#2}\centering\footnotesize}

\newcolumntype{x}[1]{>{\centering\arraybackslash}p{#1pt}}
\newcolumntype{y}[1]{>{\raggedright\arraybackslash}p{#1pt}}
\newcolumntype{z}[1]{>{\raggedleft\arraybackslash}p{#1pt}}

\makeatletter\renewcommand\paragraph{\@startsection{paragraph}{4}{\z@}
  {.5em \@plus1ex \@minus.2ex}{-.5em}{\normalfont\normalsize\bfseries}}\makeatother

\definecolor{Highlight}{HTML}{39b54a}  

\cvprfinalcopy 

\def\cvprPaperID{1917} 

\ifcvprfinal\fi

\begin{document}

\title{Geometrically Principled Connections in Graph Neural Networks}

\author{Shunwang Gong\textsuperscript{1}\thanks{Authors contributed equally.} \quad \quad Mehdi Bahri\textsuperscript{1}\footnotemark[1] \quad \quad Michael M. Bronstein\textsuperscript{1,2,3} \quad \quad Stefanos Zafeiriou\textsuperscript{1,4}\\
\textsuperscript{1}Imperial College London \quad \textsuperscript{2}Twitter\quad \textsuperscript{3}USI Lugano \quad \textsuperscript{4}FaceSoft.io\\
{\tt\small\{shunwang.gong16, m.bahri, m.bronstein, s.zafeiriou\}@imperial.ac.uk}
}

\maketitle

\begin{abstract}
    Graph convolution operators bring the advantages of deep learning to a variety of graph and mesh processing tasks previously deemed out of reach. With their continued success comes the desire to design more powerful architectures, often by adapting existing deep learning techniques to non-Euclidean data. In this paper, we argue geometry should remain the primary driving force behind innovation in the emerging field of geometric deep learning. We relate graph neural networks to widely successful computer graphics and data approximation models: radial basis functions (RBFs). We conjecture that, like RBFs, graph convolution layers would benefit from the addition of simple functions to the powerful convolution kernels. We introduce affine skip connections, a novel building block formed by combining a fully connected layer with any graph convolution operator. We experimentally demonstrate the effectiveness of our technique, and show the improved performance is the consequence of more than the increased number of parameters. Operators equipped with the affine skip connection markedly outperform their base performance on every task we evaluated, \ie, shape reconstruction, dense shape correspondence, and graph classification. We hope our simple and effective approach will serve as a solid baseline and help ease future research in graph neural networks.
\end{abstract}


\section{Introduction}
\label{sec:intro}

\begin{figure}[t!]
    \centering
    \includegraphics[width=\linewidth]{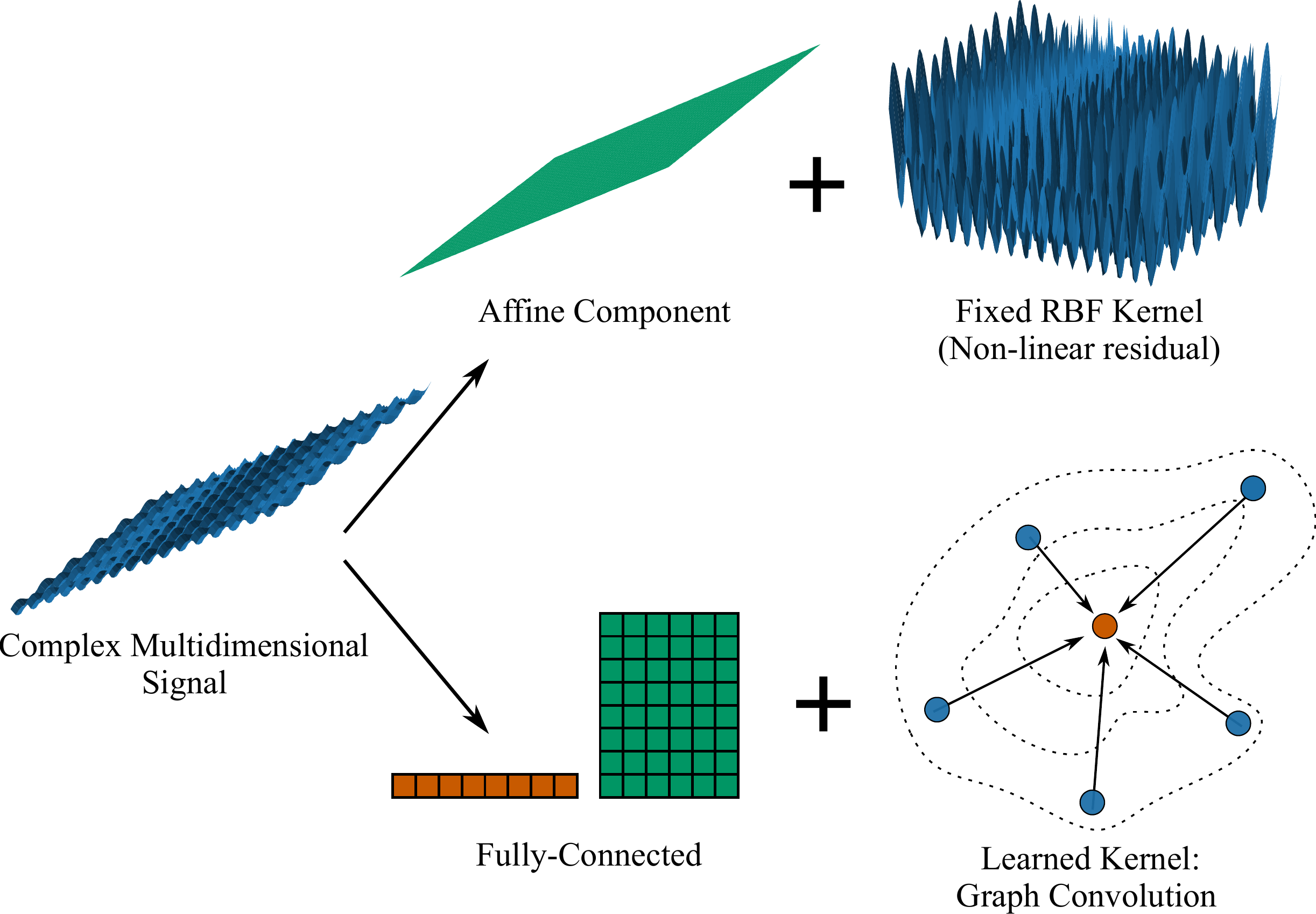}
    \caption{The comparison made in this paper between learned graph convolution kernels and RBF interpolation suggests augmenting graph convolution operators with additive affine transformations, implemented as parametric connections between layers. Our affine skip connections improve the network's ability to represent certain transformations, and enable better use of the vertex features.}
    \label{fig:cover}\vspace{-2mm}
\end{figure}

The graph formalism has established itself as the \textit{lingua franca} of non-Euclidean deep learning, as graphs provide a powerful abstraction for very general systems of interactions. In the same way that classical deep learning developed around the Convolutional Neural Networks (CNNs) and their ability to capture patterns on grids by exploiting local correlation and to build hierarchical representations by stacking multiple convolutional layers, most of the work on graph neural networks (GNNs) has focused on the formulation of convolution-like local operators on graphs. 

In computer vision and graphics, early attempts at applying deep learning to 3D shapes were based on dense voxel representations \cite{song_3d_2015} or multiple planar views \cite{Wei2015a}. These methods suffer from three main drawbacks, stemming from their \textit{extrinsic} nature: high computational cost of 3D convolutional filters, lack of invariance to rigid motions or non-rigid deformations, and loss of detail due to rasterisation. 

A more efficient way of representing 3D shapes is modeling them as surfaces (two-dimensional manifolds). 
In computer graphics and geometry processing, a popular type of efficient and accurate discretisation of surfaces are {\em meshes} or simplicial complexes (see, \eg, \cite{bronstein2006generalized, bronstein2010scale, kim2011blended, bronstein2011shape, ovsjanikov2012functional, halimi2019unsupervised, choukroun2018sparse}), which can be considered as graphs with additional structure (faces).  {\em Geometric deep learning} \cite{Bronstein2017} seeks to formulate  \textit{intrinsic} analogies of convolutions on meshes accounting for these structures. 

As a range of effective graph and mesh convolution operators are now available, the attention of the community is turning to improving the basic GNN architectures used in graph and mesh processing to match those used in computer vision. Borrowing from the existing literature, extensions of successful techniques such as residual connections \cite{He2016} and dilated convolutions \cite{yu_multi-scale_2015} have been proposed \cite{Pham2017,Rahimi2018,Xu2018}, some with major impact in accuracy \cite{Li2019}. We argue, however, that due to the particularities of meshes and to their non-Euclidean nature, geometry should be the foundation for architectural innovations in geometric deep learning.

\paragraph*{Contributions}
\label{sec:intro:contributions}

In this work, we provide a new perspective on the problem of deep learning on meshes by relating graph neural networks to Radial Basis Function (RBF) networks. Motivated by fundamental results in approximation, we introduce geometrically principled connections for graph neural networks, coined as \textit{affine skip connections}, and inspired by thin plate splines. The resulting block learns the sum of any existing graph convolution operator and an affine function, allowing the network to learn certain transformations more efficiently. 
Through extensive experiments, we show our technique is widely applicable and highly effective. We verify affine skip connections improve performance on shape reconstruction, vertex classification, and graph classification tasks. In doing so, we achieve \textit{best in class} performance on all three benchmarks. We also show the improvement in performance is significantly higher than that provided by residual connections, and verify the connections improve representation power beyond a mere increase in trainable parameters. Visualizing what affine skip connections learn further bolsters our theoretical motivation.

\mypar{Notations} Throughout the paper, matrices and vectors are denoted by upper and lowercase bold letters (\eg, $\mathbf{X}$ and  ($\mathbf{x}$), respectively.  
$\mathbf{I}$ denotes the identity matrix of compatible dimensions. The $i^{th}$ column of $\mathbf{X}$ is denoted as $\mathbf{x}_{i}$. The sets of real numbers is denoted by $\mathbb{R}$. A {\em graph} $\mathcal{G} = (\mathcal{V}, \mathcal{E})$ consists of 
{\em vertices} $\mathcal{V}=\{1,\hdots, n\}$ and {\em edges} $\mathcal{E} \subseteq \mathcal{V}\times \mathcal{V}$. The graph structure can be encoded in the {\em adjacency matrix} $\mathbf{A}$, where $a_{ij} = 1$ if $(i,j) \in \mathcal{E}$ (in which case $i$ and $j$ are said to be {\em adjacent}) and zero otherwise.  The {\em degree matrix} $\mathbf{D}$ is a diagonal matrix with elements $d_{ii} = \sum_{j=1}^n a_{ij}$. 
The {\em neighborhood} of vertex $i$, denoted by $\mathcal{N}(i) = \{j : (i,j) \in \mathcal{E} \}$, is the set of vertices adjacent to $i$. 

\section{Related work}
\label{sec:relatedwork}

\paragraph*{Graph and mesh convolutions}
\label{sec:relatedwork:gnn}

The first work on deep learning on meshes mapped local surface patches to pre-computed geodesic polar coordinates; convolution was performed by multiplying the geodesic patches by learnable filters \cite{Masci2015,Boscaini2016}. The key advantage of such an architecture is that it is intrinsic by construction, affording it invariance to isometric mesh deformations, a significant advantage when dealing with deformable shapes. 
MoNet \cite{Monti2017} generalized the approach using a local system of pseudo-coordinates $\vv{u}_{ij}$ to represent the neighborhood $\mathcal{N}(i)$ and a family of learnable weighting functions w.r.t. $\mathbf{u}$, e.g., Gaussian kernels
$
    \label{eq:monetweights}
    w_m(\vv{u}) = \exp \left( -\frac{1}{2} \trp{(\vv{u} - \bm{\mu}_m)} \bm{\Sigma}_k^{-1} (\vv{u} - \bm{\mu}_m) \right)
$ with learnable mean $\bm{\mu}_m$ and covariance $\bm{\Sigma}_m$. 
The convolution is
\begin{equation}
    \label{eq:monet}
    \vx_i^{(k)} = \sum_{m=1}^M {\theta}_m \sum_{j \in \mathcal{N}(i)} w_m(\vv{u}_{ij}) \vx_j^{(k-1)}
\end{equation} where $\vx_i^{(k-1)}$ and $\vx_i^{(k)}$  denotes the input and output features at vertex $i$, respectively, and $\bm{\theta}$ is the vector of learnable filter weights. MoNet can be seen as a Gaussian Mixture Model (GMM), and as a more general form of the Graph Attention (GAT) model \cite{Velickovic2017}. Local coordinates were re-used in the Spline Convolutional Network \cite{Fey2017}, which represents the filters in a basis of smooth spline functions. Another popular attention-based operator is FeaStNet \cite{Verma2018}, that learns a soft mapping from vertices to filter weights, and has been applied to discriminative \cite{Verma2018} and generative models \cite{Litany2017a}:
\begin{equation}
    \vx_i^{(k)} = \vv{b} + \frac{1}{|\mathcal{N}(i)|} \sum_{m=1}^M \sum_{j \in \mathcal{N}(i)} q_m(\vx_i^{(k-1)}, \vx_j^{(k-1)}) \mat{W}_m \vx_j^{(k-1)}
\end{equation} where $\mat{W}_m$ a matrix of learnable filters weights for the $m$-th filter, $q_m$ is a learned soft-assignment of neighbors to filter weights, and $\vv{b}$ the learned bias of the layer.\footnote{It is tacitly assumed here that $i \in \mathcal{N}(i)$.}

ChebNet \cite{defferrard_convolutional_2016} accelerates spectral convolutions by expanding the filters on the powers of the graph Laplacian using Chebychev polynomials. Throughout this paper, we will refer to the $n$-order expansion as ChebNet-$n$. in particular the first order expansion ChebNet-1 reads
\begin{equation}
    \label{eq:chebFirstOrder}
    \mat{X}^{(k)} = -\mat{D}^{-\frac{1}{2}} \mat{A} \mat{D}^{-\frac{1}{2}} \mat{X}^{(k-1)} \bm{\Theta}_1  + \mat{X}^{(k-1)} \bm{\Theta}_0
\end{equation} with $\mat{L} = -\mat{D}^{-\frac{1}{2}} \mat{A} \mat{D}^{-\frac{1}{2}}$ the normalised symmetric graph Laplacian, $\mat{A}$ is the graph adjacency matrix, and $\mat{D}$ is the degree matrix. In computer graphics applications, ChebNet has seen some success in mesh reconstruction and generation \cite{ranjan_generating_2018}. However, due to the fact that spectral filter coefficients are basis dependent, the spectral construction is limited to a single domain. We therefore do not evaluate the performance of ChebNet on correspondence tasks. We refer to \cite{kovnatsky2013coupled, eynard2015multimodal} for constructing compatible orthogonal bases across different domains. The Graph Convolutional Network (GCN) model \cite{Kipf2017} further simplifies (\ref{eq:chebFirstOrder}) by considering first-order polynomials with dependent coefficients, resulting in 
\begin{equation}
    \label{eq:gcn}
    \mat{X}^{(k)} = 
    \tilde{\mat{L}} \mat{X}^{(k-1)} \bm{\Theta}, 
\end{equation} 
where $\tilde{\mat{L}} = \tilde{\mat{D}}^{-\frac{1}{2}} \tilde{\mat{A}} \tilde{\mat{D}}^{-\frac{1}{2}} = \mat{I} + \mat{D}^{-\frac{1}{2}} \mat{A} \mat{D}^{-\frac{1}{2}}$. By virtue of this construction, GCN introduces self-loops. GCN is perhaps the simplest graph neural network model combining vertex-wise feature transformation (right-side multiplication by $\bm{\Theta}$) and graph propagation (left-side multiplication by $\tilde{\mat{L}}$). For this reason, it is often a popular baseline choice in the literature, but it has never applied successfully on meshes.

Recently, models based on the simple consistent enumeration of a vertex's neighbors have emerged. SpiralNet \cite{DBLP:conf/eccv/LimDCK18} enumerates the neighbors around a vertex in a spiral order and learns filters on the resulting sequence with a neural network (MLP or LSTM). The recent SpiralNet++ \cite{Gong2019} improves on the original model by enforcing a fixed order to exploit prior information about the meshes in the common case of datasets of meshes that have the same topology, \eg, \cite{Bogo:CVPR:2014,dfaust:CVPR:2017,ranjan_generating_2018}. The SpiralNet++ \cite{Gong2019} operator is written $
\vx^{(k)}_i = \gamma^{(k)} \left( \concat_{j \in S(i, M)} \vx_j^{(k-1)} \right)
$ with $\gamma^{(k)}$ an MLP, $\concat$ the concatenation, and $S(i, M)$ the spiral sequence of neighbors of $i$ of length (\ie kernel size) $M$.

Finally, we include the recently proposed Graph Isomorphism Network (GIN)  \cite{Xu2018b} with the update formula
\begin{equation}
    \vx_i^{(k)} = \gamma^{(k)}\left((1+\epsilon^{(k)}) \cdot \vx_i^{(k-1)} + \sum_{j \in \mathcal{N}(i)} \vv{x}_j^{(k-1)} \right).
\end{equation}
This model is designed for graph classification and was shown \cite{Xu2018b} to be as powerful as the Weisfeiler-Lehman graph isomorphism test. 

\paragraph*{Skip connections and GNNs}
\label{sec:relatedwork:skip}
Highway Networks \cite{srivastava2015highway, srivastava2015training} present shortcut connections with data-dependant gating functions, which are amongst the first architectures that provided a means to effectively train deep networks. However, highway networks have not demonstrated improved performance due to the fact that the layers in highway networks act as non-residual functions when a gated shortcut is "closed". Concurrent with this work, pure identity mapping \cite{He2016} made possible the training of very deep neural networks, and enabled breakthrough performance on many challenging image recognition, localization, and detection tasks. They improve gradient flow and alleviate the vanishing gradient problem. DenseNets \cite{Huang2017a} can be seen as a generalization of \cite{He2016} and connect all layers together. Early forms of skip connections in GNNs actually predate the deep learning explosion and can be traced back to the Neural Network for Graphs (NN4G) model \cite{Micheli2009}, where the input of any layer is the output of the previous layer plus a function of the vertex features \cite[section V.B]{Wu2019}.
\footnote{We refer to \cite[section 2.1]{Li2019} for a summary of subsequent approaches.}
In \cite{Li2019}, the authors propose direct graph equivalents for residual connections and dense connections, provide an extensive study of their methods, and show major improvements in the performance of the DGCNN architecture \cite{Wang2018} with very deep models.

\section{Motivation: Radial Basis Interpolation}
\label{sec:motivation}

The main motivation of this paper comes from the field of data interpolation. 
Interpolation problems appear in many machine learning and computer vision tasks. In the general setting of \textit{scattered data interpolation}, we seek a function $\hat{f}$ whose outputs $\hat{f}(\vx_i)$ on a set of scattered data points $\vx_i$ equals matching observations $\vv{y}_i$, \ie, $\forall i, f(\vx_i) = \vv{y}_i$. In the presence of noise, one typically solves an approximation problem potentially involving regularization, \ie
\begin{equation}
\min_f \sum_i d(\hat{f}(\vx_i), \vv{y}_i) + \lambda L(\hat{f}),
\end{equation} where $d$ measures the adequation of the model $\hat{f}$ to the observations, $\lambda$ is a regularization weight, and $L$ encourages some chosen properties of the model. For the sake of the discussion, we take $d(\vx, \vv{y}) = ||\vx-\vv{y}||$. In computer graphics, surface reconstruction and deformation (\eg for registration \cite{Chen2017}) can be phrased as interpolation problems.

In this section, we draw connections between graph convolutional networks and a classical popular choice of interpolants: Radial Basis Functions (RBFs).

\paragraph*{Radial basis functions}
\label{sec:motivation:rbfnets}

An RBF is a function of the form
$
\vx \mapsto \phi(||\vx-\vv{c}_i||)
$,
with $||.||$ a norm, and $\vv{c}_i$ some pre-defined centers. By construction, the value of an RBF only depends on the distance from the centers. While an RBF function's input is scalar, the function can be vector-valued.

In interpolation problems, the centers are chosen to be the data points ($\vv{c}_i = \vx_i$) and the interpolant is defined as a weighted sum of radial basis functions centered at each $\vx_i$:
\begin{equation}
    \hat{f}(\vx) = \sum_{i=1}^N w_i \phi(||\vx-\vx_i||).
\end{equation}
Interpolation assumes equality, so the problem boils down to solving the linear system
$
    \bm{\Phi} \vv{w}_i = \vv{b}_j
$,
with $\bm{\Phi}_{j,i} = \phi(||\vx_i - \vx_j||)$ the matrix of the RBF kernel (note that the diagonal is $\phi(\vv{0}) \; \forall i$). The kernel matrix encodes the relationships between the points, as measured by the kernel.

Relaxing the equality constraints can be necessary, in which case we solve the system in the least squares sense with additional regularization. We will develop this point further to introduce our proposed affine skip connections.

\paragraph*{Relations to GNNs}
\label{sec:motivation:gcn}

An RBF function can be seen as a simple kind of one layer neural network with RBF activations centered around every points (\ie an RBF network \cite{Broomhead1988,Moody1989}). The connection to graph neural networks is very clear: while the RBF matrix encodes the relationships and defines a point's neighborhood radially around the point, graph neural networks rely on the graph connectivity to hard-code spatial relationships. In the case of meshes, this encoding is all-the-more relevant, as a notion of distance is provided either by the ambient space (the graph is embedded) or directly on the Riemannian manifold. The latter relates to the RBFs with geodesic distance of \cite{Rhee2011}.

Most GNNs used on meshes fall into the message passing framework \cite{Gilmer2017}:
\begin{align}
\begin{split}
    \label{eq:mpnn}
     & \vx_i^{(k)}  = \\
     & \gamma^{(k)}\left( \vx_i^{(k-1)}, \displaystyle \mathop{\square}_{j \in \mathcal{N}(i)} \phi^{(k)}\left(\vx_i^{(k-1)}, \vx_j^{(k-1)}, \vv{e}_{ij}^{(k-1)}\right) \right),
\end{split}
\end{align} where $\square$ denotes a differentiable permutation-invariant function, (\eg $\max$ or $\sum$), $\phi$ a differentiable kernel function, $\gamma$ is an MLP, and $\mathbf{x}_i$ and $\vv{e}_{ij}$ are features associated with vertex $i$ and edge $(i,j)$, respectively. This equation defines a compactly supported, and possibly \textit{non-linear}, function around the vertex. For the MoNet equation (\ref{eq:monet}) the connection to RBFs is direct. Contrary to RBFs, the filters of modern GNNs do not have to be radial. In fact, anisotropic filters \cite{Boscaini2016,bouritsas2019neural} have been shown to perform better than isotropic ones \cite{Masci2015,ranjan_generating_2018}. The other major differences are:
\begin{enumerate}
    \item The filters are learned functions, not pre-defined; this allows for better inductive learning and task-specificity
    \item The filters apply to any vertex and edge features
    \item Some operators support self-loops, but $\mathrm{diag}(\bm{\Phi}) = \phi(\vv{0})$ irrespective of the features $\vv{x}_i$
\end{enumerate}

We note that the compact support of (\ref{eq:mpnn}) is a design decision: early GNNs built on the graph Fourier transform lacked compactly-supported filters \cite{Henaff2015}. In RBF interpolation, global support is sometimes desired as it is a necessary condition for maximal \textit{fairness} of the interpolated surfaces (\ie maximally smooth), but also induces computational complexity and numerical challenges as the dense kernel matrices grow and become ill-conditioned \cite{Anjyo:2014:SDI:2614028.2615425}. This motivated the development of fast methods to fit locally supported RBFs \cite{Beatson1999}. In \cite{Henaff2015} the authors argue compactly-supported kernels are desirable in graph neural networks for computational efficiency, and to promote learning local patterns. This especially justified for meshes, for which the graph structure is very sparse. Additionally, stacking convolutional layers is known to increase the receptive field, including in graph neural networks \cite{Wu2019}. The composition of locally supported filters can therefore yield globally supported mappings.

\paragraph*{RBFs and polynomials}
\label{sec:motivation:polys}

A common practice with RBFs is to add low-order polynomial terms to the interpolant:
\begin{equation}
    \label{eq:rbfpoly}
    \hat{f}(\vx) = \sum_{i=1}^N w_i \phi(||\vx-\vx_i||) + P(\vx).
\end{equation}

The practical motivation is to ensure polynomial mappings of some order can be represented exactly and to avoid unwanted oscillations when approximating flat functions, \eg affine transformations of an image should be exactly affine. One can show \cite{Anjyo:2014:SDI:2614028.2615425} this is equivalent to ensuring the RBF weights lie in the null space of the polynomial basis, also known as the \textit{vanishing moments condition}.

However, polynomials appear organically when the RBF kernel is derived to be optimal for a chosen roughness measure, typically expressed in terms of the integral of a squared differential operator $D$ (below in one dimension):
\begin{equation}
    \label{eq:squareddiffoperator}
    ||Df||^2 = \int |Df(x)|^2dx,
\end{equation} \eg, $D = \frac{d^2}{dx^2}$.
In other words, when the kernel is sought to be optimal for a given \textit{regularization functional}. Differential operators are very naturally expressed on meshes in terms of finite difference approximations. In this case, we identify $D$ with its corresponding stencil matrix. The interpolation problem becomes the minimization of (\ref{eq:squareddiffoperator}) subject to the interpolation constraints.

It can be shown \cite{Anjyo:2014:SDI:2614028.2615425} that for such problems the RBF kernel is the Green's function of the squared differential operator, and that for an operator of order $m$, polynomials of order $m-1$ span the null space. Therefore, the complete solution space is the direct sum\footnote{Hence the vanishing moment condition.} of the space of polynomials of order $m-1$ (the null space of the operator) and the space spanned by the RBF kernel basis\footnote{This result comes from phrasing the problem as regularization in a Reproducing Kernel Hilbert Space. To keep the discussion short in this manuscript, we refer the reader to relevant resources such as \cite[Section 7]{Anjyo:2014:SDI:2614028.2615425}.}.

\mypar{Thin Plate Splines (TPS)} \label{sec:motivation:tps}
An important special case is the RBF interpolant for a surface $z(\vx), \; \vx = \trp{[x \quad y]}$ that minimizes the bending energy $\int \int \frac{\partial^2 f}{\partial x^2} + \frac{\partial^2 f}{\partial x \partial y} + \frac{\partial^2 f}{\partial y^2} dxdy = ||\Delta^2 f||$. The solution is the well-known biharmonic spline, or \textit{thin plate spline}, $\phi(r) = r^2 \log r, \quad r = ||\vx - \vx_i||$, with a polynomial of degree 1 (\ie an affine function)
\begin{equation}
    \label{eq:tps}
    \hat{f}(\vx) = \sum_i w_i \phi(||\vx - \vx_i||) + \mat{A} \vx + \vv{b}.
\end{equation}
Generalizations to higher dimensions yield \textit{polyharmonic splines}. These splines maximize the surface \textit{fairness}. From (\ref{eq:tps}) it is also clear \textit{the polynomial doesn't depend on the structure of the point set} and is common for all points.

\section{Geometrically Principled Connections}
\label{sec:block}

In Section \ref{sec:motivation}, we highlighted key similarities and differences between continuous RBFs and discrete graph convolution kernels. We then exposed how adding low-order polynomials to RBFs kernels is both beneficial to enable efficient fitting of flat functions, and deeply connected to regularization of the learned functions, and noted the polynomial component does not depend on spatial relationships. Based on these observations, we conjecture that graph convolution operators could, too, benefit from the addition of a low-order polynomial to ensure they can represent flat functions exactly, and learn functions of a vertex's features independently from its neighbours. We introduce a simple block that achieves both goals.

\begin{figure}
    \centering
    \includegraphics[width=.7\linewidth]{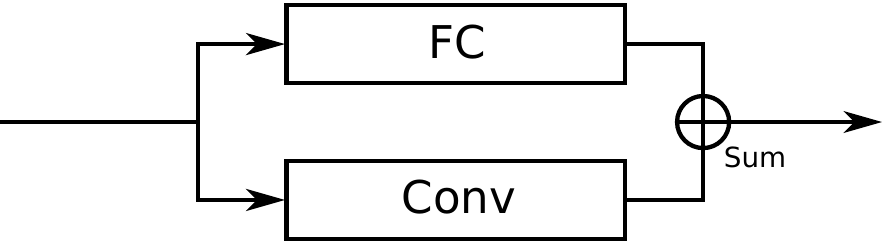}
    \caption{Our block learns the sum of one graph convolution and a shortcut equipped with an affine transformation.}
    \label{fig:block_diagram}\vspace{-2mm}
\end{figure}

\label{sec:block:description}
Inspired by equation (\ref{eq:tps}), we propose to augment a \textit{generic} graph convolution operator with \textit{affine skip connections}, \ie, inter-layer connections with an affine transformation implemented as a fully connected layer. The output of the block is the sum of the two paths, as shown in Figure \ref{fig:block_diagram}.

Our block is designed to allow the fully connected layer to learn an affine transformation of the current feature map, and let the convolution learn a residue from a vertex's neighbors. For message passing, we obtain:
\begin{align}
\begin{split}
    \label{eq:mpnnskip}
    & \vx_i^{(k)}  = \\
    & \gamma^{(k)}\left( \vx_i^{(k-1)}, \displaystyle\mathop{\square}_{j \in \mathcal{N}(i)} \phi^{(k)}(\vx_i^{(k-1)}, \vx_j^{(k-1)}, \vv{e}_{i,j}^{(k-1)}) \right) \\
    & + \mat{A}^{(k)} \vx_i^{(k-1)} + \vv{b}^{(k)} .
\end{split}
\end{align}

The fully connected layer could be replaced by an MLP to obtain \textit{polynomial connections}, however, we argue the stacking of several layers creates sufficiently complex mappings by composition to not require deeper sub-networks in each block: a balance must be found between expressiveness and model complexity. Additionally, the analogy with TPS appears well-motivated for signals defined on surfaces. \textbf{As a matter of notation, we refer to our block based on operator Conv with affine skip connections as \emph{Aff}-Conv.}

In equations (\ref{eq:rbfpoly}), (\ref{eq:tps}) and (\ref{eq:mpnnskip}), the polynomial part does not depend on a vertex's neighbors, but solely on the feature at that vertex. This is similar to PointNet \cite{Qi2017a} that learns a shared MLP on all points with no structural prior. In our block, the geometric information is readily encoded in the graph, while the linear layer is applied to all vertices independently, thus learning indirectly from the other points regardless of their proximity.

\mypar{Residual blocks with projections} In \cite[Eq. (2)]{He2016}, the authors introduced a variation of residual blocks with a projection implemented as a linear layer. Their motivation is to handle different input and output sizes. We acknowledge the contribution of residual connections and will demonstrate our block provides the same benefits and more for GNNs.

\section{Experimental evaluation}
\label{sec:experiments}

Our experiments are designed to highlight different properties of affine skip connections when combined. We present the individual experiments, then draw conclusions based on their entirety. All implementation details (model architecture, optimizers, losses, \etc), and details about the datasets (number of samples, training/test split) are provided in Appendix A of the supplementary material.

\subsection{Experimental design}

\begin{table*}[t]
\vspace{-.5em}
\tablestyle{3pt}{1.05}
\begin{tabular}{x{60}|x{60}x{26}x{26}|x{60}x{26}x{26}|x{60}x{26}x{26}c}
 &
\multicolumn{3}{c|}{M=4} & 
\multicolumn{3}{c|}{M=9} & 
\multicolumn{3}{c}{M=14} & \\
method &
mean error &
median &
\# param &
mean error &
median &
\# param &
mean error &
median &
\# param &\\ 
\shline
ChebNet$^\dagger$ & 0.659 $\pm$ 0.783 & 0.391 & 92.5k & 4.329 $\pm$ 3.591 & 3.453 & 154.9k & 4.348 $\pm$ 3.587 & 3.469 & 217.3k \\
ChebNet & \bd{0.520} $\pm$ \bd{0.655} & \bd{0.294} & 92.5k & \bd{0.438} $\pm$ \bd{0.562} & \bd{0.244} & 154.9k & \bd{0.407} $\pm$ \bd{0.523} & \bd{0.227} & 217.3k \\
\emph{Res}-ChebNet & 0.531 $\pm$ 0.668 & 0.299 & 92.5k & 0.444 $\pm$ 0.570 & 0.275 & 154.9k & 0.412 $\pm$ 0.530 & 0.229 & 217.3k \\
\hline
SpiralNet++$^\dagger$ & \bd{0.554} $\pm$ \bd{0.674} & \bd{0.320} & 92.5k & 0.430 $\pm$ 0.542 & 0.239 & 154.9k & 0.385 $\pm$ 0.491 & 0.214 & 217.3k \\
SpiralNet++ & 0.578 $\pm$ 0.705 & 0.333 & 92.5k & \bd{0.426} $\pm$ \bd{0.538} & \bd{0.238} & 154.9k & \bd{0.383} $\pm$ \bd{0.489} & \bd{0.212} & 217.3k \\
\emph{Res}-SpiralNet++ & 0.575 $\pm$ 0.703 & 0.331 & 92.5k & 0.432 $\pm$ 0.541 & 0.243 & 154.9k & 0.395 $\pm$ 0.496 & 0.223 & 217.3k \\
\hline
FeaStNet & 0.599 $\pm$ 0.730 & 0.342 & 93.8k & 0.524 $\pm$ 0.646 & 0.297 & 157.9k & 0.488 $\pm$ 0.599 & 0.279 & 221.9k \\
FeaStNet+ & 0.587 $\pm$ 0.723 & 0.333 & 106.6k & 0.517 $\pm$ 0.635 & 0.292 & 170.7k & 0.477 $\pm$ 0.594 & 0.268 & 234.8k \\
\emph{Res}-FeaStNet & 0.565 $\pm$ 0.701 & 0.314 & 93.8k & 0.483 $\pm$ 0.602 & 0.266 & 157.9k & 0.441 $\pm$ 0.554 & 0.279 & 221.9k \\
\emph{Aff}-FeaStNet & \bd{0.543} $\pm$ \bd{0.676} & \bd{0.303} & 106.3k & \bd{0.470} $\pm$ \bd{0.585} & \bd{0.261} & 170.4k & \bd{0.431} $\pm$ \bd{0.543} & \bd{0.237} & 234.4k \\
\hline
MoNet & 0.671 $\pm$ 0.760 & 0.450 & 92.7k & 0.528 $\pm$ 0.604 & 0.354 & 155.4k & 0.480 $\pm$ 0.551 & 0.321 & 218.1k \\
MoNet+ & 0.627 $\pm$ 0.693 & 0.429 & 105.2k & 0.528 $\pm$ 0.587 & 0.366 & 167.9k & 0.480 $\pm$ 0.540 & 0.329 & 230.6k \\
\emph{Res}-MoNet & 0.540 $\pm$ 0.612 & 0.335 & 92.7k & 0.426 $\pm$ 0.479 & 0.271 & 155.4k & 0.374 $\pm$ 0.417 & 0.238 & 218.1k \\
\emph{Aff}-MoNet & \bd{0.499} $\pm$ \bd{0.579} & \bd{0.298} & 105.2k & \bd{0.406} $\pm$ \bd{0.455} & \bd{0.251} & 167.9k & \bd{0.347} $\pm$ \bd{0.386} & \bd{0.218} & 230.5k \\
\end{tabular}
\vspace{1mm}
\caption{\textbf{3D shape reconstruction} experiments results in the CoMA \cite{ranjan_generating_2018} dataset. Errors are in millimeters. All the experiments were ran with the same network architecture. We show the results of each operator for different kernel sizes (\text{i.e.}, \# of weight matrices). \textit{Aff-} denotes the operators equipped with the proposed affine skip connections, \textit{Res-} denotes the operators with standard residual connections, and $\dagger$ indicates we remove the separate weight for the center vertex.}
\label{tab:shape_reconstruction_results}
\vspace{-3mm}
\end{table*}

\mypar{Mesh reconstruction} The task is to reconstruct meshes with an auto-encoder architecture, and relates the most to interpolation. To validate the proposed approach, we firstly show the performance of attention-based models, MoNet and FeaStNet, on shape reconstruction on CoMA\cite{ranjan_generating_2018} for different values of $M$. For a kernel size of $M$, we compare the vanilla operators (MoNet, FeaStNet), the blocks with residual skip connections (\textit{Res}-MoNet, \textit{Res}-FeaStNet), the blocks with affine skip connections (\textit{Aff}-MoNet, \textit{Aff}-FeaStNet), and the vanilla operators with kernel size $M+1$ (MoNet+, FeaStNet+)\footnote{Increasing the kernel size by 1, and adding an affine skip connection lead to the same number of weight matrices.}. We evaluated kernel sizes 4, 9, and 14. We report the mean Euclidean vertex error and its standard deviation, and the median Euclidean error. Results with SplineCNN \cite{fey2018splinecnn} are shown in Appendix B of the supplementary material.

\mypar{Mesh correspondence} The experimental setting is mesh correspondence, \ie, registration formulated as classification. We compare MoNet, FeaStNet and their respective blocks on the FAUST \cite{Bogo:CVPR:2014} dataset. We purposefully do not include SpiralNet++ and ChebNet on this problem: the connectivity of FAUST is {\em fixed} and vertices are in correspondence already. These methods assume a fixed topology and therefore have an unfair advantage. We report the percentage of correct correspondences as a function of the geodesic error.

\mypar{Mesh correspondence with GCN} The GCN \cite{Kipf2017} model is arguably the most popular graph convolution operator, and has been widely applied to problems on generic graphs thanks to its simplicity. However, its performance degrades quickly on meshes, which makes the entry bar higher for prototyping graph-based approaches in 3D vision. We investigate whether affine skip connections can improve the performance of GCN, and by how much. We choose the 3D shape correspondence task, in order to allow for comparison with the other models already included in this study. As detailed in the supplementary material, the network used in this experiment is relatively deep, with three convolution layers. In \cite[Appendix B]{Kipf2017} the authors add residual connections to GCNs deeper than two layers to alleviate vanishing gradients. In order to prove affine skip connections have a geometric meaning, we must eliminate the possibility that better performance comes solely from improved gradient flow. We include in this study a GCN block with vanilla residual connections (\textit{Res}-GCN), in order to isolate the gradient flow improvements from the geometric improvements. Overall, we compare vanilla GCN, \textit{Res}-GCN, and our \textit{Aff}-GCN.

\begin{figure}[t]
\newcommand{\h}{0.5\linewidth}
\includegraphics[width=.99\linewidth]{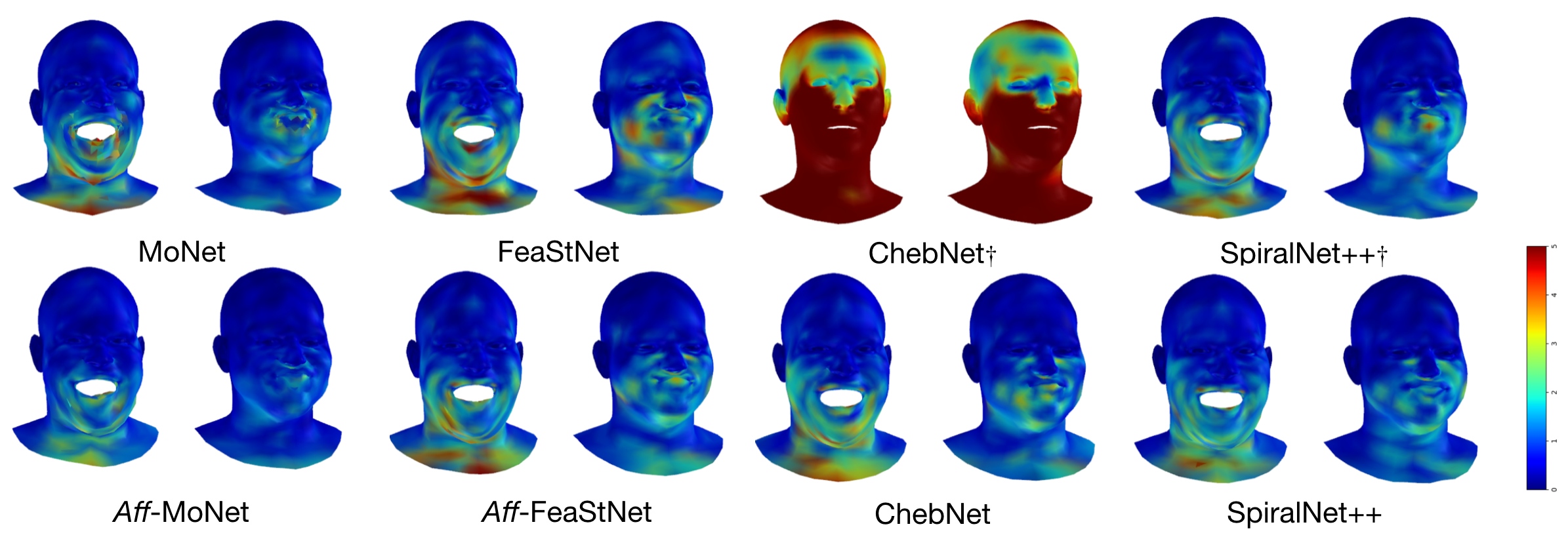} \\
\makebox[\h][c]{\footnotesize (a) Addition (MoNet, FeaStNet)}\hfill%
\makebox[\h][c]{\footnotesize (b) Ablation (ChebNet, SpiralNet++)}\\\vspace{-0.5\baselineskip}
\caption{\textbf{Sample reconstructions:} addition of affine skip connections and ablation of the center vertex weights. 
}
\label{fig:recon-samples}\vspace{-2mm}
\end{figure}

\begin{figure}[t]
    \centering
    \includegraphics[width=.73\linewidth]{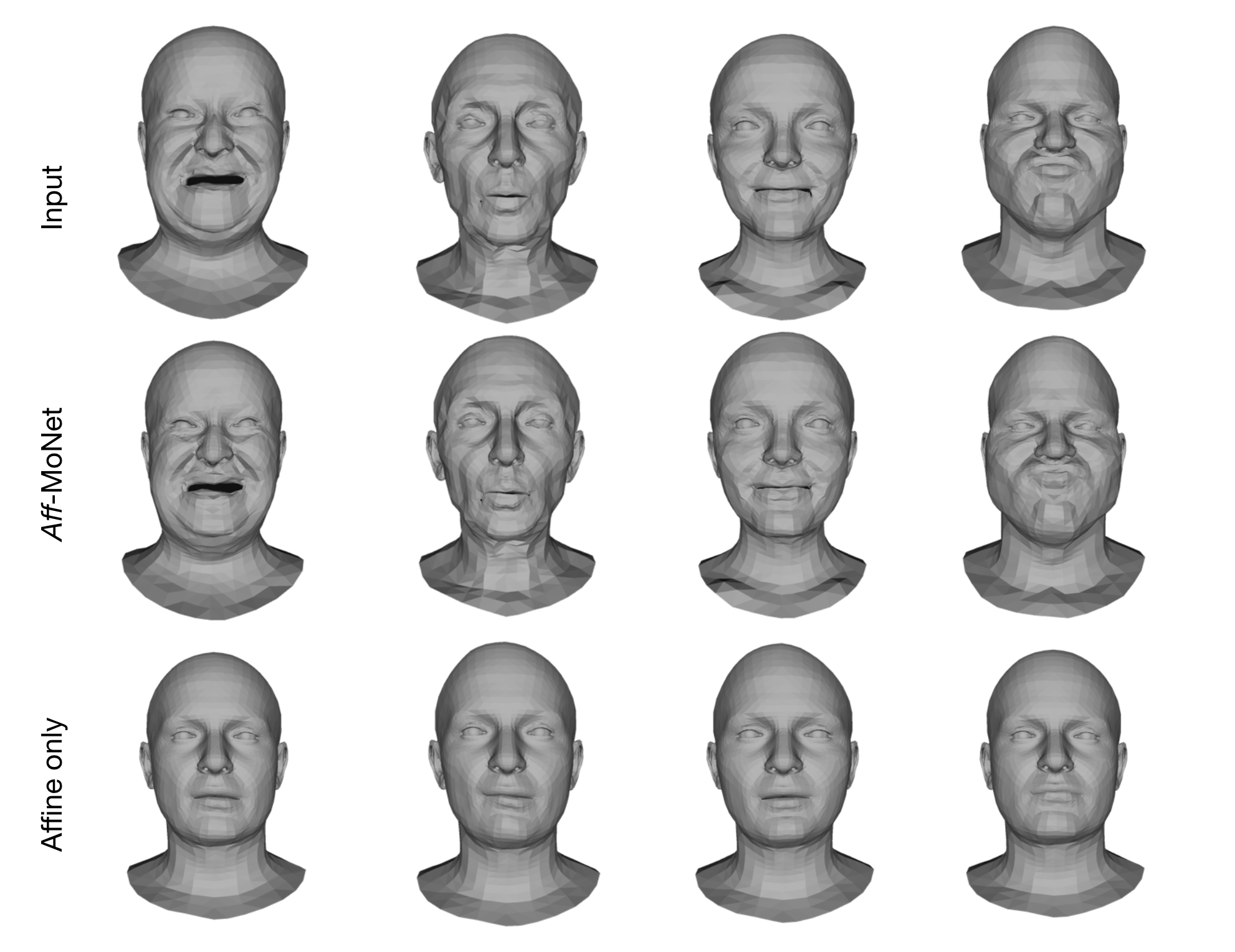}
    \caption{
    Example reconstructed faces obtained by passing samples (\textbf{top}) through a trained autoencoder built on the \textit{Aff}-MoNet block. The \textbf{middle} row shows reconstructions produced by the full autoencoder. The \textbf{bottom} row shows the result of passing through the affine skip connections only in the decoder at inference. The connections learn a smooth component common to the samples - across identities and expressions, as expected from the motivation.
    }
    \label{fig:monetlinearonly}\vspace{-2mm}
\end{figure}

\begin{figure*}[t]
    \centering
    \includegraphics[width=0.99\linewidth]{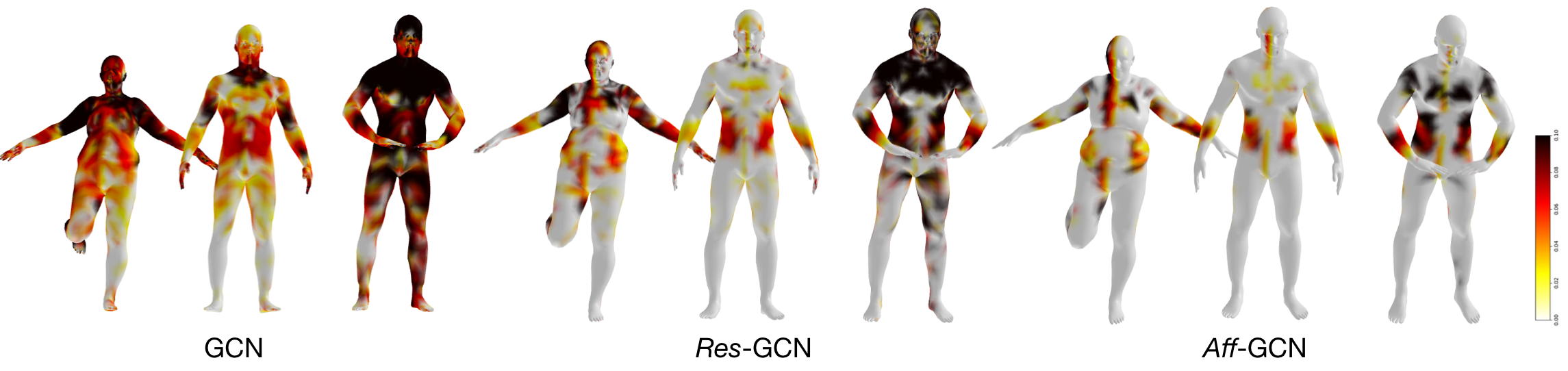}\vspace{-2mm}
    \caption{\textbf{Shape correspondence} experiments on the FAUST humans dataset. Per-vertex heatmap of the geodesic error for three variants of the GCN operator.  Geodesic error is measured according to
 \cite{Kim2011}.} 
    \label{corresp:a}\vspace{-3mm}
\end{figure*}

\newcommand{\h}{0.45\linewidth}
\begin{figure}

\begin{minipage}{.5\linewidth}
\centering
\subfloat[MoNet \& FeaStNet]{\label{corresp:b}\includegraphics[width=\linewidth]{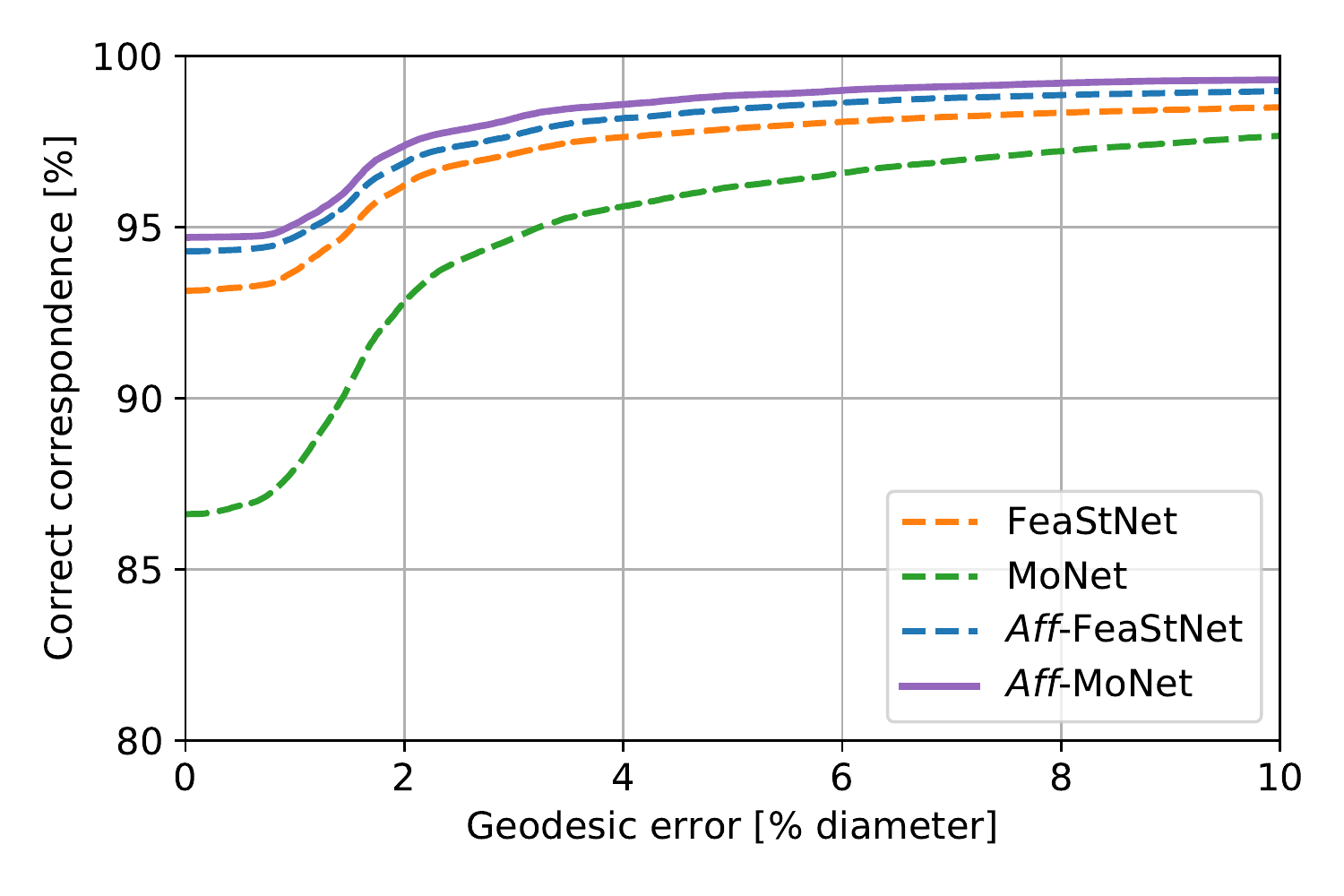}}
\end{minipage}%
\begin{minipage}{.5\linewidth}
\centering
\subfloat[GCN]{\label{corresp:c}\includegraphics[width=\linewidth]{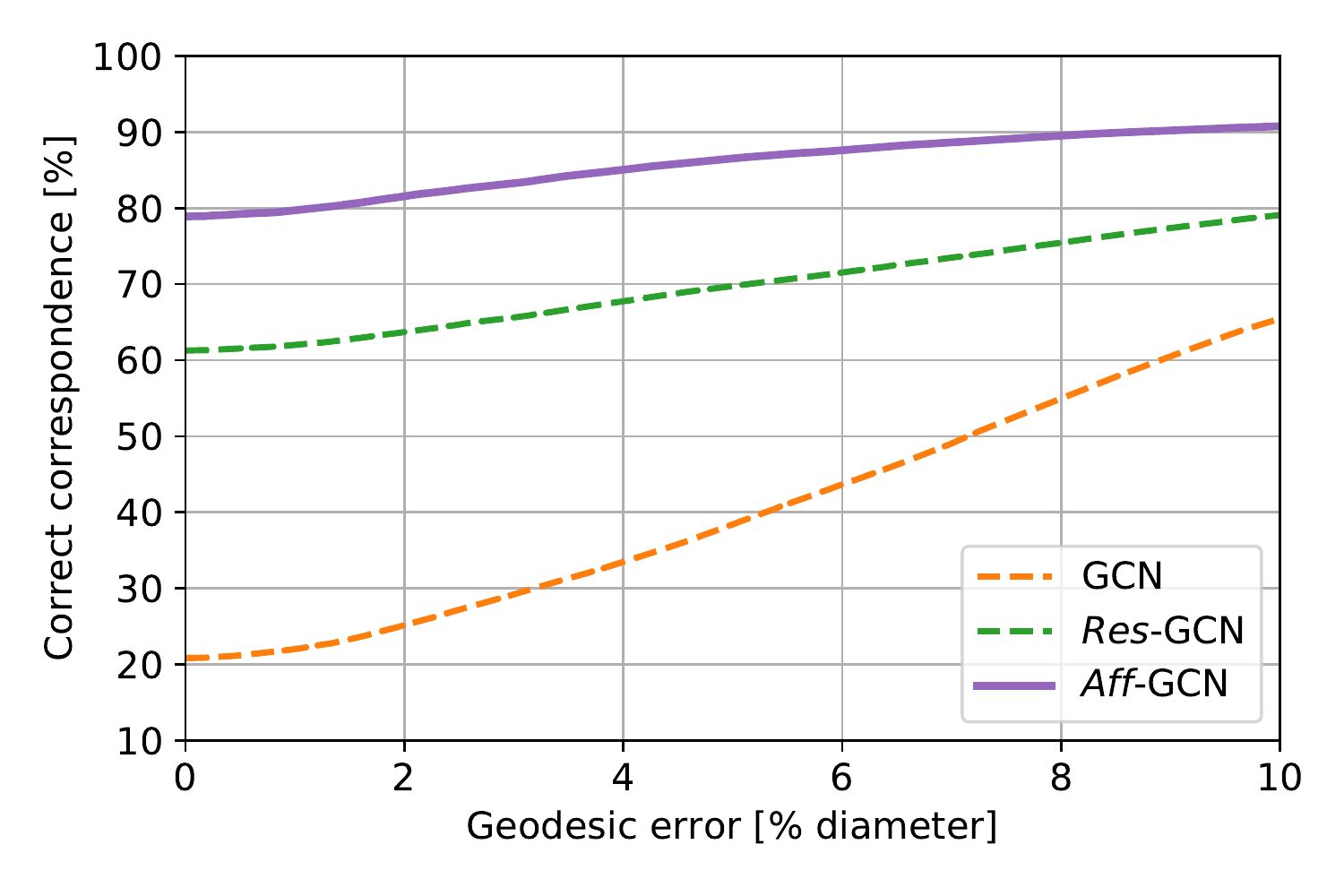}}
\end{minipage}\par\medskip
\vspace{-2mm}
\caption{\textbf{Shape correspondence} accuracy: the $x$ axis displays the geodesic error in \% of the mesh diameter, and the $y$ axis shows the percentage of correspondences that lie within a given radius around the correct vertex. All experiments were ran with the same architecture. \textit{Aff}-GCN only has \textbf{1\%} more parameters than GCN.
}
\label{fig:corresp}\vspace{-2mm}
\end{figure}

\mypar{Graph classification} We compare MoNet, FeaStNet, and their respective residual and affine skip connection blocks on graph classification on Superpixel MNIST \cite{Monti2017, Fey2017}.  The Superpixel MNIST dataset used in \cite{Monti2017} and \cite{Fey2017} represents the MNIST images as graphs. 
We use 75 vertices per image. All models use a kernel size of 25. We include GIN (built with a 2-layer MLP) for the similarity of its update rule with our block, in the GIN-0 ($\epsilon = 0$) variant for its superior performance  as observed in \cite{Xu2018b}. We compare GIN with GCN, \textit{Res-GCN}, and \textit{Aff}-GCN. Here, graph connectivity is \textbf{not} fixed. We report the classification accuracy.

\mypar{Ablation: separate weights for the centre vertex} To show the inclusion of the center vertex is necessary, we perform an ablation study of ChebNet, and SpiralNet++ on shape reconstruction on CoMA. From equation (\ref{eq:chebFirstOrder}), we see the zero order term $\mat{X}\bm{\Theta}_0$ is an affine function of the vertex features. We remove it from the expansion of ChebNet-$(M+1)$ to obtain ChebNet-$M^{\dagger}$:
$
\label{eq:chebAblaMOrder}
\mat{X}^{(k)} = \mat{L}^{(M+1)} \mat{X}^{(k-1)} \bm{\Theta}_{M+1} + \ldots + \mat{L} \mat{X}^{(k-1)}  \bm{\Theta}_1.
$
Both models have identical numbers of weight matrices, but ChebNet-$M$ learns from the vertices alone at order $0$. For SpiralNet++, the center vertex is the first in the sequence $\{vertex||neighbors\}$. We rotate the filter (\ie move it one step down the spiral) to remove the weight on the center vertex while keeping the same sequence length. We obtain SpiralNet++$^{\dagger}$. The number of weight matrices is constant. All models have kernel size 9.

\mypar{Ablation: self-loops \vs affine skip connections} We also compare FeaStNet with and without self-loops (FeaStNet$^\dagger$), and the matching blocks, on all experiments.

\subsection{Results and discussion}
Based on the evidence collected, we draw conclusions about specific properties of our affine skip connections.

\mypar{Parameter specificity} The results of varying the kernel size on shape reconstruction can be found in Table \ref{tab:shape_reconstruction_results} along with the corresponding number of parameters for control.
Increasing the kernel size by 1 (MoNet+, FeaStNet+) provides only a minor increase in performance, \eg, for $M=9$ and $M=14$, MoNet and MoNet+ have the same mean Euclidean error and the median error of MoNet with $M=9$ actually \textit{increases} by $\mathbf{3.4\%}$. In contrast, the affine skip connections always drastically reduce the reconstruction error, for the same number of additional parameters. In particular, the mean Euclidean error of MoNet decreased by $\mathbf{25.6\%}$ for $M = 4$, and by $\mathbf{23.1\%}$ for $M = 9$. We conclude our affine skip connections have a specific different role and augment the representational power of the networks beyond simply increasing the number of parameters. Our block with MoNet achieves the new \textit{state of the art} performance on this task.

\mypar{What do affine skip connections learn?} In Figure \ref{fig:monetlinearonly}, we observe the linear layers in the connections learned information common to all shapes. This result strengthens our analogy with the polynomial terms in RBF interpolation: the coefficients of the polynomial function are learned from all data points and shared among them. In one dimension, this can be pictured as learning the trend of a curve. Our visualizations are consistent with this interpretation.

\begin{table}[t]
\tablestyle{2.5pt}{1.2}
\begin{tabular}{l|x{30}x{30}x{50}x{30}}
 Method & \multicolumn{2}{c}{Acc. (\%)} & Kernel Size & \# Param \\ 
\shline
         GIN-0        &  \multicolumn{2}{c}{57.75} & - & 25k \\
         \hline
        GCN     &  \multicolumn{2}{c}{31.21} & - & 15.9k\\
        \emph{Res}-GCN &  \multicolumn{2}{c}{42.32} & - & 15.9k\\
        \emph{Aff}-GCN   &  \multicolumn{2}{c}{\textbf{58.96}} & - & 22.1k\\
        \hline
        FeaStNet    &  \multicolumn{2}{c}{11.35} &  25 & 166k\\
        \emph{Res}-FeaStNet    &  \multicolumn{2}{c}{58.09} &  25 & 166k\\
        \emph{Aff}-FeaStNet  &  \multicolumn{2}{c}{\textbf{59.50}} & 25 &  172k\\
        \hline
        \textit{Pseudo-Coord.}       & \textit{Degree} & \textit{Position} & - & -\\
        MoNet       & 53.10 &  96.57  & 25 & 164k\\
        \emph{Res}-MoNet       & 53.75 &  96.82  & 25 & 164k\\
        \emph{Aff}-MoNet     & \textbf{72.00} &  \textbf{97.14}  & 25 &  170k\\
\end{tabular}\vspace{2mm}
\caption{
\textbf{Classification accuracy} of different operators and blocks on the Superpixel MNIST dataset with 75 superpixels.
For MoNet, we report performance using pseudo-coordinates computed from the vertex positions, or from the connectivity only (vertex degrees).
}
\label{tab:exp_mnist}\vspace{-2mm}
\end{table}

\begin{table*}[t]\vspace{-3mm}
\centering
\subfloat[\textbf{Shape Reconstruction}: ]{
\tablestyle{2.5pt}{1.1}\begin{tabular}{c|x{55}x{25}x{30}x{22}}
 \scriptsize method & mean error &  median & \# param & M\\
\shline
 \scriptsize FeaStNet & 0.524 $\pm$ 0.646 & 0.297 & 157.9k & 9 \\
 \scriptsize \emph{Aff}-FeaStNet & \bd{0.470} $\pm$ \bd{0.585} & 0.261 & 170.4k & 9 \\\hline
 \scriptsize FeaStNet$^\dagger$ & 0.519 $\pm$ 0.634 & 0.297 & 157.9k & 9 \\
 \scriptsize \emph{Aff}-FeaStNet$^\dagger$ & \bd{0.463} $\pm$ \bd{0.577} & 0.256 & 170.4k & 9 \\
\end{tabular}}\hspace{1mm}
\subfloat[\textbf{Correspondence}]{
\tablestyle{2.5pt}{1.05}\begin{tabular}{x{35}x{35}x{22}}
   acc. (\%) & \# param. & M\\
\shline
  93.14 & 1.91M & 10\\
  \bd{94.29} & 1.92M & 10\\\hline
  93.72 & 1.91M & 10\\
  \bd{94.36} & 1.92M & 10\\
\end{tabular}}\hspace{3mm}
\subfloat[\textbf{Classification}]{
\tablestyle{2.5pt}{1.05}\begin{tabular}{x{30}x{30}x{22}}
   acc. (\%) & \# param. & M\\
\shline
  11.35 & 166k & 25\\
  \bd{59.50} & 172k & 25\\\hline
  11.35 & 166k & 25\\
  \bd{60.07} & 172k & 25\\
\end{tabular}}\vspace{-1mm}
\caption{\textbf{Ablations:} \textbf{affine skip connection \vs self-loop}. We show the performances of FeaStNet under the settings of with and without self-loop (denoted with $^\dagger$) and with and without affine skip connections regarding the tasks of shape reconstruction on CoMA, shape correspondence on FAUST, and classification on MNIST with 75 superpixels. $M$ denotes the kernel size (\ie \# weight matrices). For correspondence, test accuracy is the ratio of the correct correspondence prediction at geodesic error $0$.}
\label{tab:ablations_feastnet}\vspace{-2mm}
\end{table*}

\mypar{Vertex-level representations} We report the mesh correspondence accuracy as a function of the geodesic error for FeaStNet, MoNet, and the blocks in Figure \ref{corresp:b}. We observe consistent performance improvements for both operators. The performance difference is remarkable for MoNet: for a geodesic error of $0$, the accuracy improved from $86.61\%$ to $94.69\%$. \textit{Aff}-MoNet is the new \textit{state of the art} performance on this problem\footnote{Excluding methods that learn on a fixed topology.}. We conclude affine skip connections improve vertex-level representations.

\mypar{Laplacian smoothing and comparison to residuals} We show the performance of GCN and its residual and affine blocks in Figure \ref{corresp:c}. The accuracy of vanilla GCN is only around 20\%. We can hypothesize this is due to the equivalence of GCN with Laplacian smoothing \cite{Li2018} - blurring the features of neighboring vertices and losing specificity - or to the vanishing gradient problem. Our block outperforms vanilla residuals by a large margin: the classification rate of \textit{Aff}-GCN is nearly 79\% while \textit{Res}-GCN only reaches 61.27\%. Visually (Figure \ref{corresp:a}), \textit{Res}-GCN provides marked improvements over GCN, and \textit{Aff}-GCN offers another major step-up. A similar trend is seen in Table \ref{tab:shape_reconstruction_results} and Table \ref{tab:exp_mnist}. In \cite{He2016} the authors observed a minor performance increase between vanilla residuals and residual connections with projection, that they attributed to the higher number of parameters. The differences we observe are not consistent with such marginal improvements. This shows \textit{not only} our approach provides all the benefits of residuals in solving the vanishing gradient problem, it achieves \textit{more} on \textit{geometric data}, and that the improvements are not solely due to more trainable parameters or improved gradient flow. In particular, with affine skip connections, Eq. 4 of \cite{Li2018} becomes $\sigma(\mathbf{\tilde{L}}\mathbf{H}^{(l)}\bm{\Theta}^{(l)} + \mathbf{H}^{(l)}\mathbf{W}^{(l)})$, with $\mathbf{\tilde{L}}$ the augmented symmetric Laplacian, and $\mathbf{W}^{(l)}$ the parameters of the affine skip connection. Thus, the \textit{Aff}-GCN block is no longer equivalent to Laplacian smoothing.

\mypar{Discriminative power} Our results on Superpixel MNIST are presented in Table \ref{tab:exp_mnist}. Our affine skip connections improve the classification rate across the board. GCN with affine skip connections outperform GIN-0 by over 1 percentage point, with $12\%$ fewer trainable parameters. This result shows \textit{Aff}-GCN offers competitive performance with a smaller model, and suggests the augmented operator is significantly more discriminative than GCN. Assuming the terminology of \cite{Xu2018b}, FeaStNet employs a \textit{mean} aggregation function, a choice known \cite{Xu2018b} to significantly limit the discriminative power of GNNs and which could explain its very low accuracy in spite of its large (166k) number of parameters. In contrast, \textit{Aff}-FeaStNet is competitive with \textit{Aff}-GCN and outperforms GIN-0. As GIN is designed to be as powerful of the WL test, these observations suggest affine skip connections improve the discriminative power of graph convolution operators. As a result, \textit{Aff}-MoNet outperformed the current \textit{state of the art}, for coordinate-based and degree-based pseudo-coordinates.

\mypar{Role of the center vertex} As seen in the first six rows of Table \ref{tab:shape_reconstruction_results}, the performance of the models is higher with weights for the center vertex, especially for ChebNet. Note the comparison is at identical numbers of parameters. Figure \ref{fig:recon-samples} provides sample ablation and addition results. This shows convolution operators need to learn from the center vertices. We found that removing self-loops in FeaStNet actually \textit{increased} the performance for both the vanilla and the block operators. Table \ref{tab:ablations_feastnet} shows results on all experiments. The affine skip connection consistently improved the performance of models regardless of the self-loops. We conclude graph convolution operators should be able to learn specifically from the center vertex of a neighborhood, independently from its neighbors. A similar observation was made in \cite{Xu2018b} where independent parameters for the center vertex are shown to be required for graph convolution operators to be as discriminative as the WL test.

\section{Conclusion}

By relating graph neural networks to the theory of radial basis functions, we introduce \textit{geometrically principled connections} that are both easily implemented, applicable to a broad range of convolution operators and graph or mesh learning problems, and highly effective. We show our method extends beyond surface reconstruction and registration, and can dramatically improve performance on graph classification with arbitrary connectivity. Our MoNet block achieves state of the art performance and is more robust to topological variations than sequence (SpiralNet++) or spectrum-based (ChebNet) operators. We further demonstrate our blocks improve on vanilla residual connections for graph neural networks. We believe our approach is therefore interesting to the broader community. Future work should study whether affine skip connections have regularization effects on the smoothness of the learned convolution kernels.

\paragraph*{Acknowledgements} 
S.G. and M.M.B. were supported by ERC Consolidator grant No. 724228 (LEMAN). M.B. was supported by a PhD Scholarship from Imperial College London, a Qualcomm Innovation Fellowship, and acknowledges the support of Amazon through the AWS Cloud Credits for Research program. S.Z. was partially funded by the EPSRC Fellowship DEFORM: Large Scale Shape Analysis of Deformable Models of Humans (EP/S010203/1) and an Amazon AWS Machine Learning Research Award.

{\small
\bibliographystyle{ieee_fullname}
\bibliography{main.bib}
}

\clearpage
\appendix

\section*{Supplementary Material}
\begin{figure*}[t]
    \centering
    \includegraphics[width=.97\linewidth]{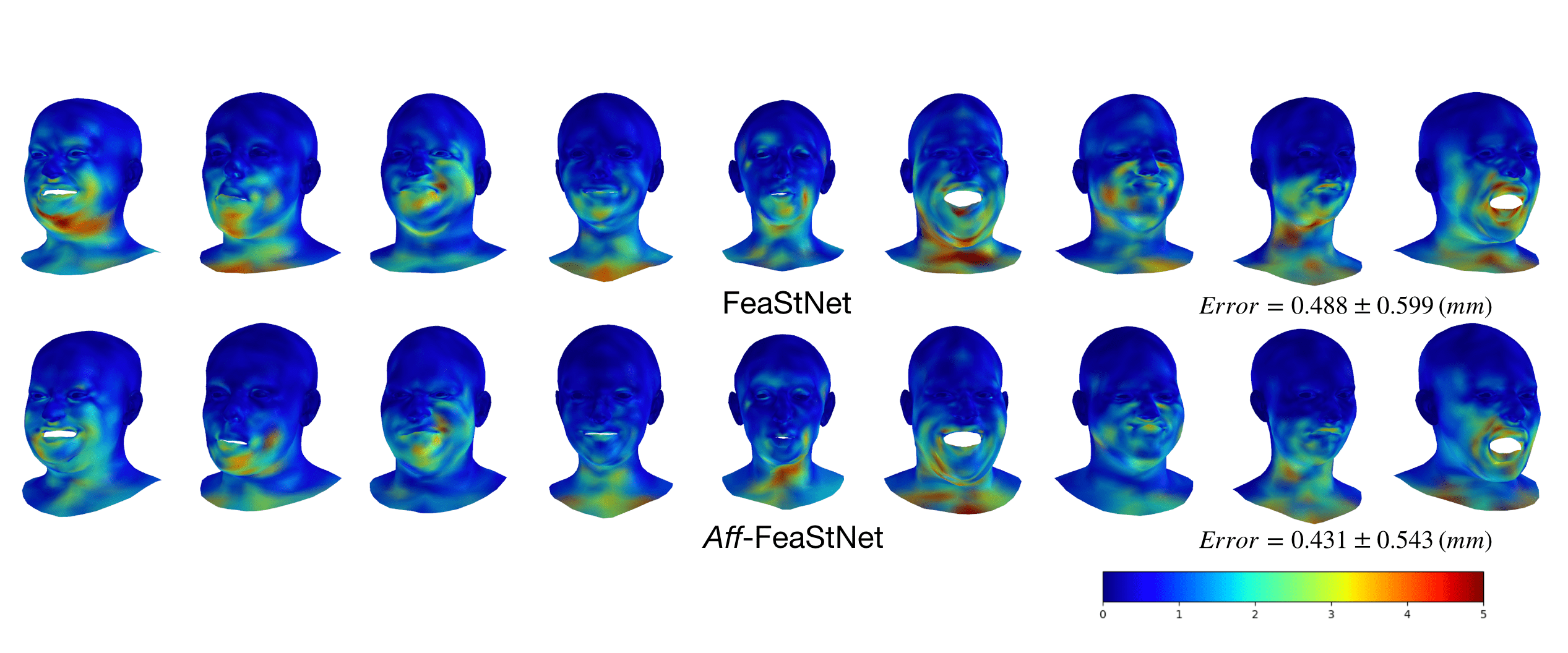} \\
    \includegraphics[width=.97\linewidth]{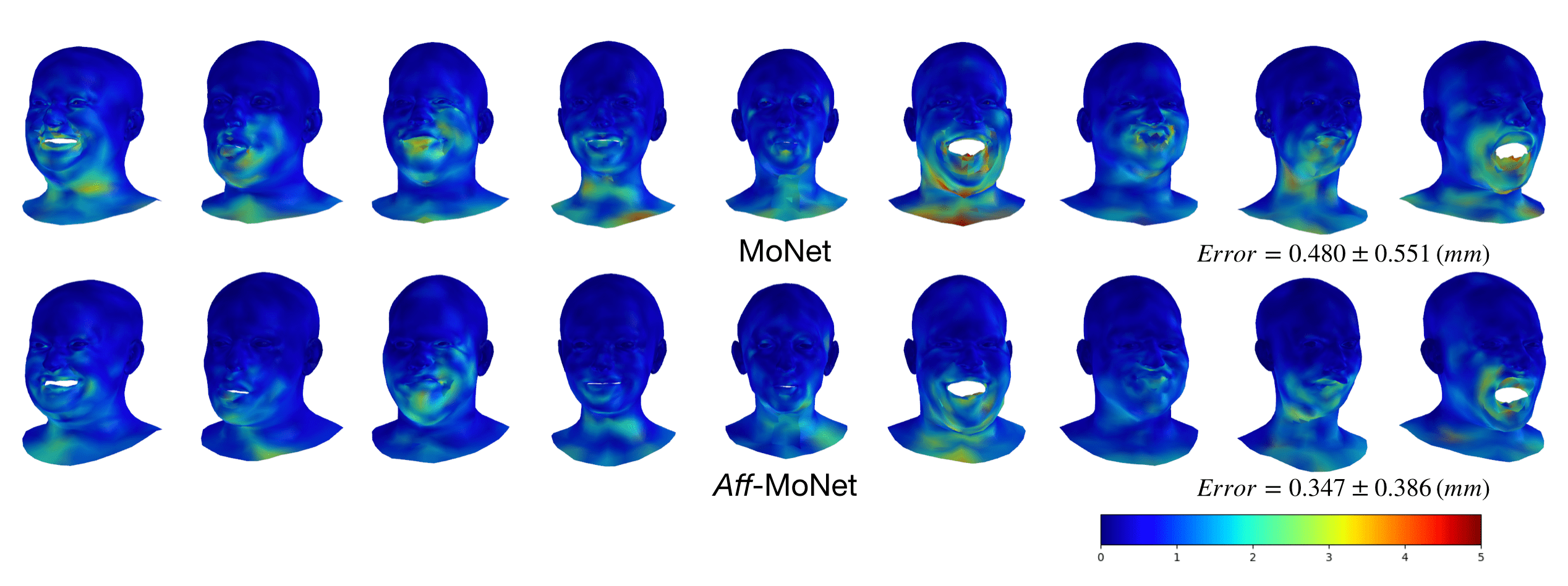}
    \caption{
    Pointwise error (\textit{Euclidean distance} from groundtruth) of the reconstructions by FeaStNet \cite{Verma2018} and MoNet \cite{Monti2017} (both with and without affine skip connections) on the CoMA \cite{ranjan_generating_2018} test dataset. The reported errors (bottom-right corner of each row) represent the per-point mean error and its standard deviation. For visualization clarity, the error values are saturated at 5 millimeters. Hot colors represent large errors.
    }
    \label{fig:reconstruction_feast_monet}
    \vspace{-2mm}
\end{figure*}

\begin{figure}[t]
    \centering
    \includegraphics[width=.90\linewidth]{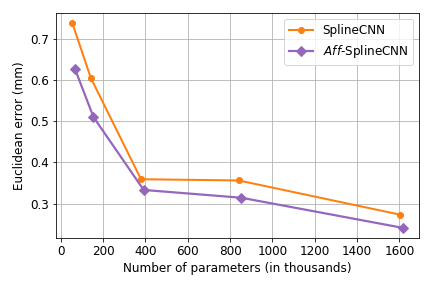}
    \caption{
The pointwise mean euclidean error of SplineCNN and \textit{Aff}-SplineCNN for shape reconstruction experiments on the CoMA \cite{ranjan_generating_2018} dataset. 
    }
    \label{fig:spline}
    \vspace{-10mm}
\end{figure}

\begin{figure*}[t]
    \centering
    \includegraphics[width=.97\linewidth]{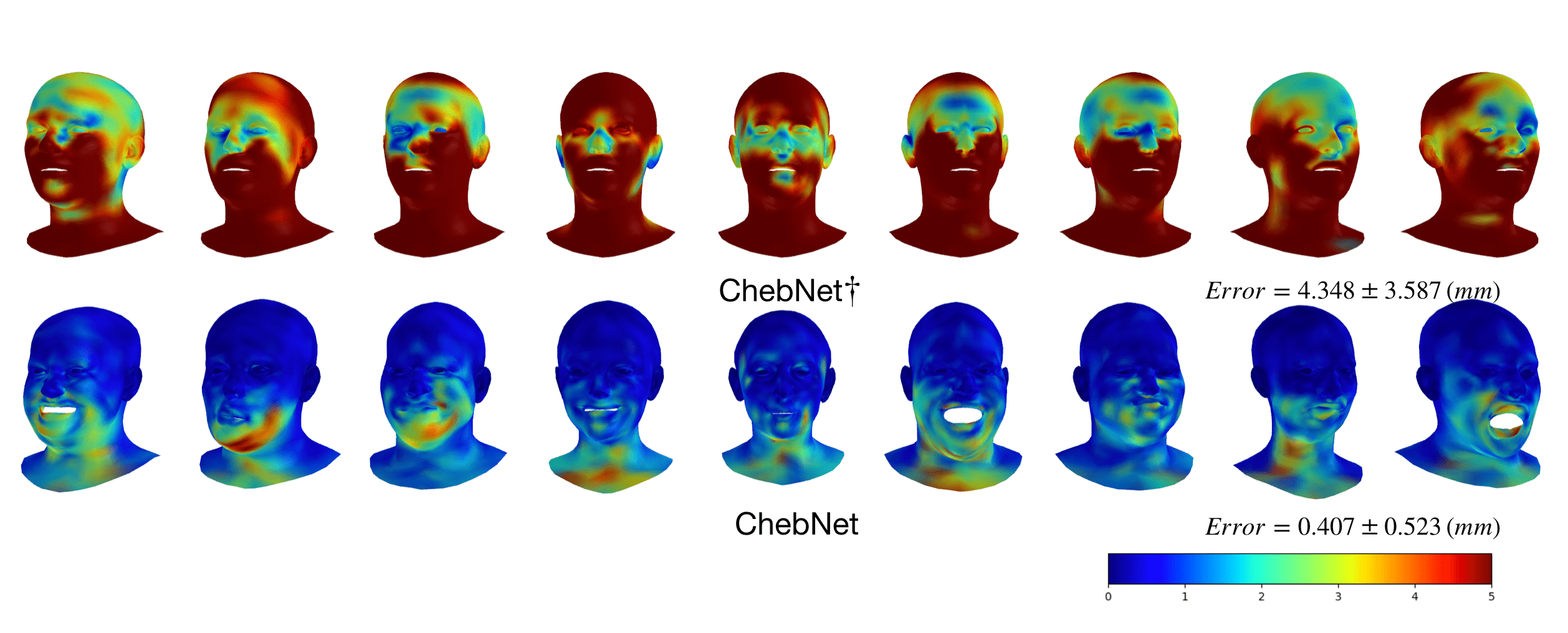}\\
    \includegraphics[width=.97\linewidth]{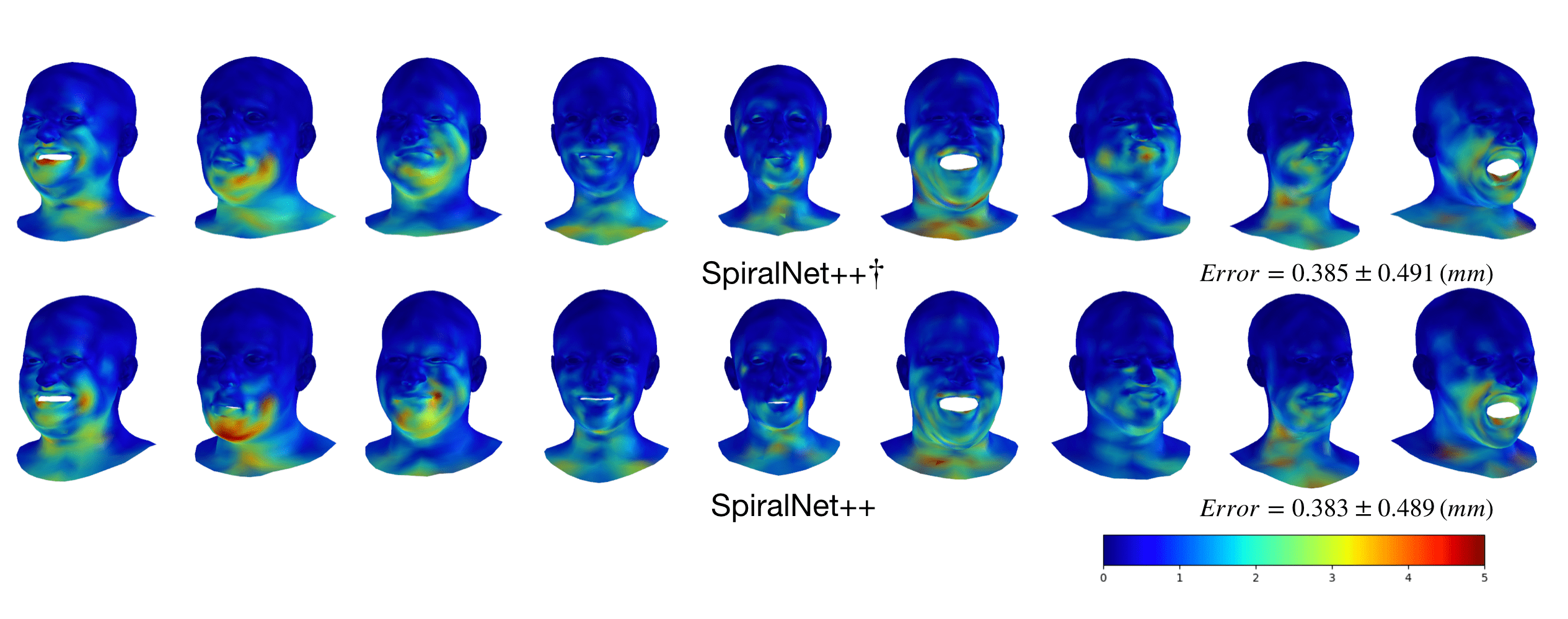}
    \caption{
    Pointwise error (\textit{Euclidean distance} from groundtruth) of the reconstructions by ChebNet \cite{defferrard_convolutional_2016} and SpiralNet++ \cite{Gong2019} (\textit{ablation study}) on the CoMA \cite{ranjan_generating_2018} test dataset. We reformulated ChebNet and SpiralNet++ to \textit{remove the separate weight of the center vertex} while keeping the same number of weight matrices, denoted as $\dagger$. The detailed formulas are explained in Section 5.1 of the paper. The reported errors (bottom-right corner of each row) represent the per-point mean error and its standard deviation. For visualization clarity, the error values are saturated at 5 millimeters. Hot colors represent large errors.
    }
    \label{fig:reconstruction_cheb_spiral}
\end{figure*}

\begin{figure*}[t]
    \centering
    \includegraphics[width=.99\linewidth]{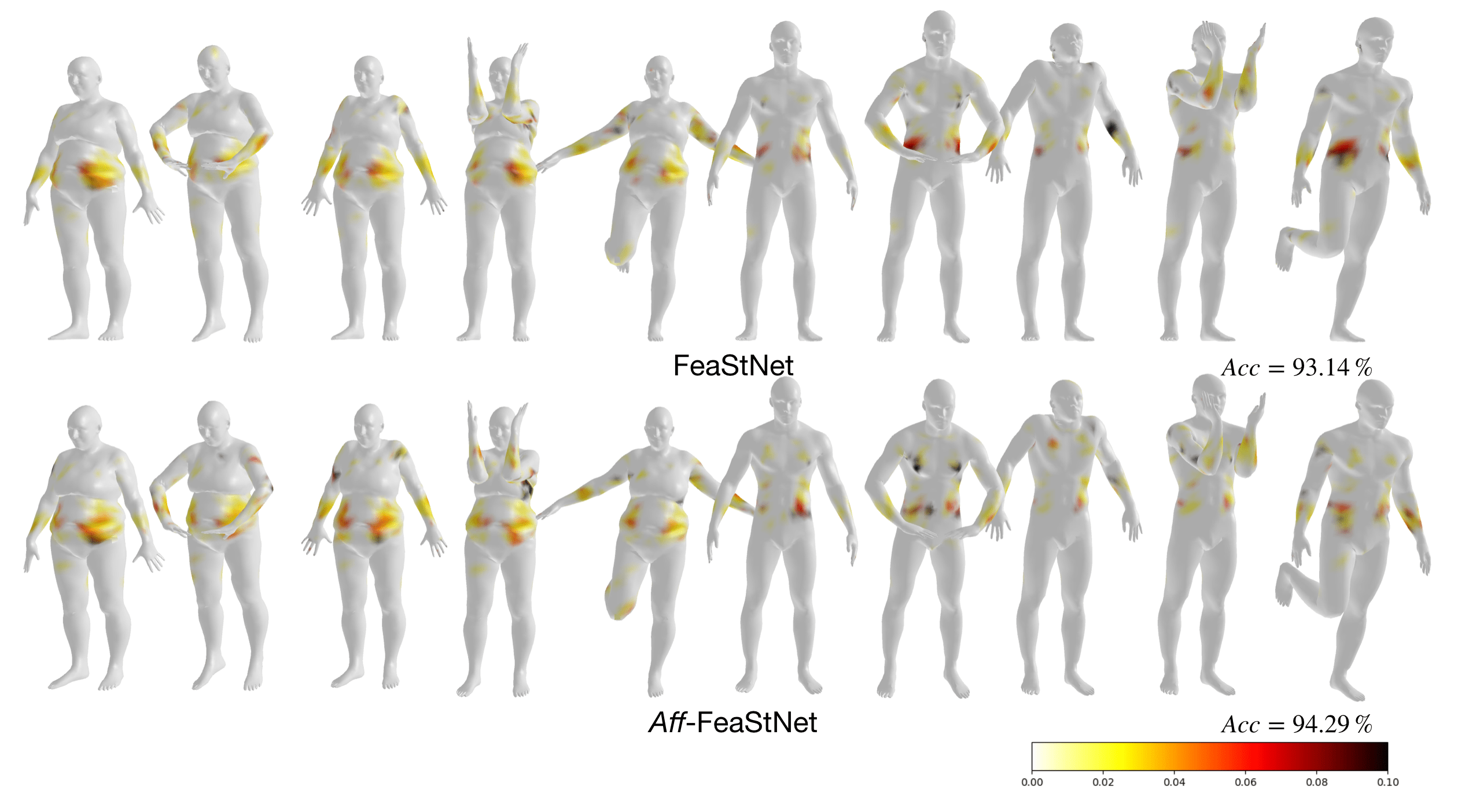}   
    \includegraphics[width=.99\linewidth]{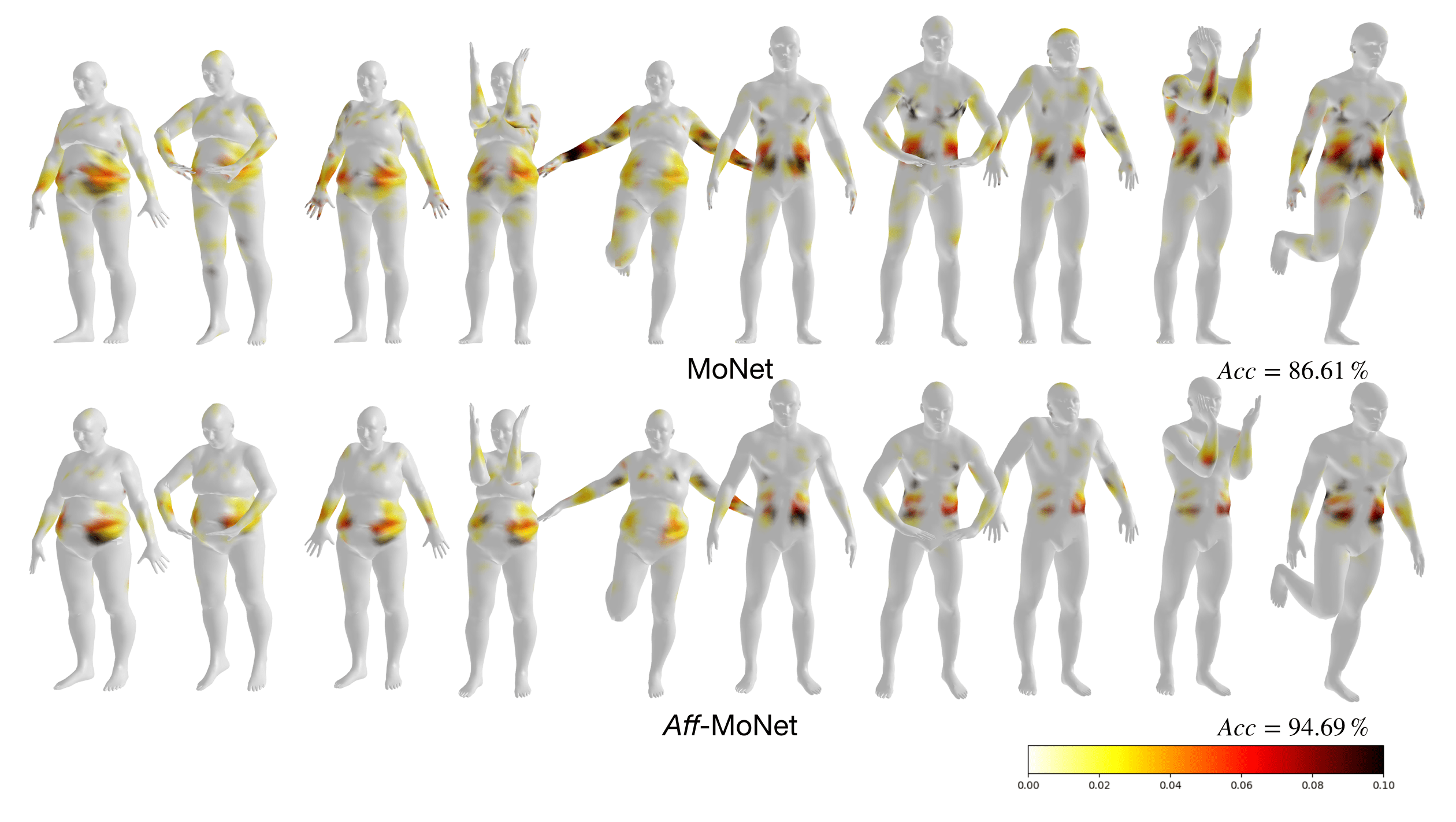}    
    \caption{
Pointwise error (\textit{geodesic distance} from groundtruth) of FeaStNet \cite{Verma2018} and MoNet \cite{Monti2017} (both with and without affine skip connections) on the FAUST \cite{Bogo:CVPR:2014} humans dataset. The reported accuracy values (bottom-right corner of each row) represent the percentage of correct correspondence at geodesic error 0. For visualization clarity, the error values are saturated at 10\% of the geodesic diameter. Darker colors represent large errors.}
    \label{fig:correspondence_feast_monet}
    \vspace{10mm}
\end{figure*}

\begin{figure*}[t]
    \centering
    \includegraphics[width=.99\linewidth]{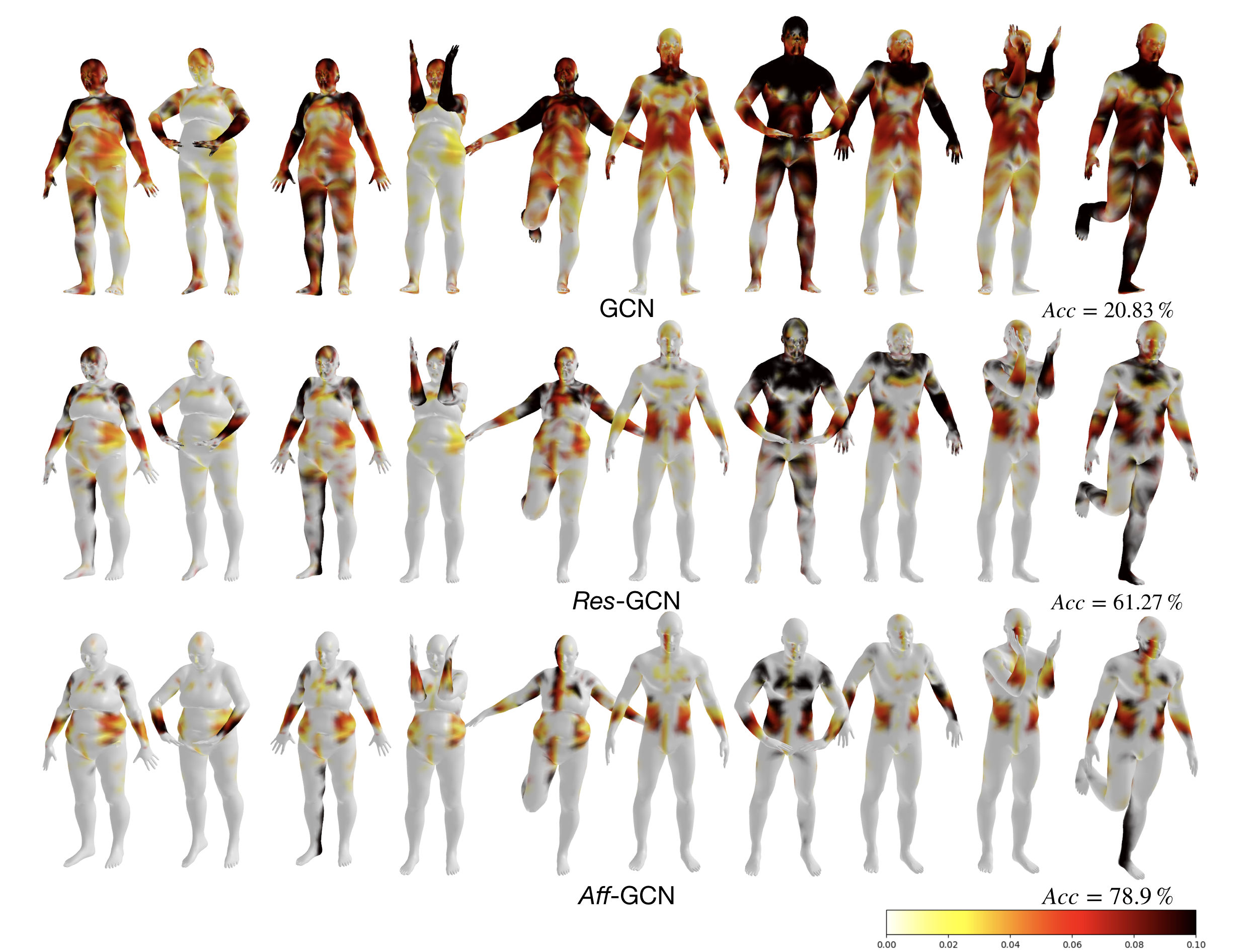}
    \caption{
Pointwise error (\textit{geodesic distance} from groundtruth) of vanilla GCN \cite{Kipf2017}, \textit{Res}-GCN and \textit{Aff}-GCN on the FAUST  \cite{Bogo:CVPR:2014} humans dataset. \textit{Aff}-GCN replaces the residual connections of \textit{Res}-GCN to the proposed affine skip connections. The rest are the same. The reported accuracy values (bottom-right corner of each row) represent the percentage of correspondence at geodesic error 0. For visualization clarity, the error values are saturated at 10\% of the geodesic diameter. Darker colors represent large errors.
    }
    \label{fig:correspondence_gcn}
    \vspace{10mm}
\end{figure*}

This supplementary material provides further details that is
not be included in the main text: Section \ref{sec:setup} provides implementation details on the experiments used in Section 5 of the paper, and Section \ref{sec:spline} further describes the results obtained by SplineCNN \cite{fey2018splinecnn} with and without the proposed affine skip connections on the task of shape reconstruction.  Figures~\ref{fig:reconstruction_feast_monet} and \ref{fig:reconstruction_cheb_spiral} show the faces reconstructed by autoencoders built with each convolution operator presented in Table 1 of the paper, at kernel size 14. Figures~\ref{fig:correspondence_feast_monet} and \ref{fig:correspondence_gcn} show the visualization of shapes colored by the pointwise geodesic error of different methods on the FAUST \cite{Bogo:CVPR:2014} humans dataset.

\section{Implementation Details}
\label{sec:setup}

For all experiments we initialize all trainable weight parameters with Glorot initialization \cite{Glorot2010} and biases with constant value 0. The only exception is FeaStNet \cite{Verma2018}, for which weight parameters (e.g. W, $\mu$, c) are drawn from $\mathcal{N}(0, 0.1)$. The vertex features fed to the models are the raw 3D Cartesian coordinates (for the CoMA \cite{ranjan_generating_2018} and FAUST datasets) or the 1D superpixel intensity (for the Superpixel MNIST dataset \cite{Monti2017}). The pseudo-coordinates used in MoNet \cite{Monti2017} and SplineCNN \cite{fey2018splinecnn} are the pre-computed relative Cartesian coordinates of connected nodes. Note that in Superpixel MNIST classification experiments, we compared the performance of MoNet using pseudo-coordinates computed from relative Cartesian coordinates which considering vertex positions as well as globally normalized degree of target nodes for the sake of the fairness. All experiments are ran on a single NVIDIA RTX 2080 Ti.

\mypar{Shape reconstruction} We perform experiments on the CoMA dataset \cite{ranjan_generating_2018}. We follow the interpolation experimental setting in \cite{ranjan_generating_2018}, the dataset is split in training and test sets with a ratio of $9:1$. We normalize the input data by subtracting the mean and dividing by the standard deviation \textit{obtained on the training set} and we de-normalize the output before visualization. We quantitatively evaluate models with the pointwise Euclidean error (we report the mean, standard deviation, and median values) and the visualizations for qualitative evaluation.

The experimental setting is identical to \cite{Gong2019}. The network architecture is 3 $\times$ \{Conv(32)$\rightarrow$ Pool(4)\} $\rightarrow$ \{Conv(64) $\rightarrow$ Pool(4)\} $\rightarrow$ FC(16) for the encoder, and a symmetrical decoder with one additional Conv(3) output to reconstruct 3D coordinates, with ELU activations after each convolutional layer except on the output layer that has no activate. We used the same downsampling and upsampling approach introduced in \cite{ranjan_generating_2018}.  Models are trained with Adam \cite{kingma2014adam} for 300 epochs with an initial learning rate of 0.001 and a learning rate decay of 0.99 per epoch, minimizing the $\ell_1$ vertex-wise loss. The batch size is 32.

\mypar{Mesh correspondence} We perform experiments on the FAUST dataset \cite{Bogo:CVPR:2014}, containing 10 scanned human shapes in 10 different poses, resulting in a total of 100 non-watertight meshes with 6,890 nodes each. The first 80 subjects in FAUST were used for training and the remaining 20 subjects for testing, following \cite{Monti2017}. Correspondence quality is measured according to the Princeton benchmark protocol \cite{kim2011blended}, counting the percentage of derived correspondences that lie within a geodesic radius $r$ around the correct node. 

We use the single scale architecture of \cite{Verma2018} with an added dropout layer. We obtain the architecture Lin($16$)$\rightarrow$Conv($32$)$\rightarrow$Conv($64$)$\rightarrow$Conv($128$)$\rightarrow$Lin($256$)\\
$\rightarrow$Dropout($0.5$)$\rightarrow$Lin($6890$), where Lin($o$) denotes a $1\times1$ convolution layer that produces $o$ output features per node. We use ELU non-linear activation functions after each Conv layer, and after the first Lin layer. We use a softmax activation for the last layer. Models are trained with the standard cross-entropy loss for $1000$ epochs. We use the Adam optimizer with an initial learning rate of $0.001$ for MoNet (with and without affine skip connections) and GCN (vanilla, \textit{Res} and \textit{Aff}), and an initial learning rate of $0.01$ for FeaStNet (with and without affine skip connections). We decay the learning rate by a factor of $0.99$ every epoch for MoNet (with and without affine skip connections) and GCN (vanilla, \textit{Res} and \textit{Aff}), and a factor of $0.5$ every $100$ epochs for FeaStNet (with and without affine skip connections). We use a batch size of 1. Note that for \textit{Res}-GCN, we use \textbf{zero-padding shortcuts for mismatched dimensions}.

\mypar{Superpixel MNIST classification} Experiments are conducted on the Superpixel MNIST dataset introduced in \cite{Monti2017}, where MNIST images are represented as graphs with \textit{different} connectivity, each containing $75$ vertices. The dataset is split into training and testing sets of 60k and 10k samples respectively. 

Our architecture is similar to the one introduced in \cite{Monti2017} with three convolutional layers, and reads Conv($32$)$\rightarrow$Pool($4$)$\rightarrow$Conv($64$)$\rightarrow$Pool($4$)$\rightarrow$Conv($64$)$\rightarrow$AvgP\\
$\rightarrow$FC($128$)$\rightarrow$Dropout($0.5$)$\rightarrow$FC($10$). Pool($4$) is based on the Graclus graph coarsening approach, downsampling graphs by approximately a factor of $4$. AvgP denotes a readout layer that averages features in the node dimension. As for the nonlinearity, ELU activation functions are used after each layer except for the last layer that uses softmax. We train networks using the Adam optimizer for $500$ epochs, with an initial learning rate of $0.001$ and learning rate decay of $0.5$ after every $30$ epochs. We minimize the cross-entropy loss. The batch size is 64 and we use $\ell_2$ regularization with a weight of $0.0001$. For each GIN-$0$ \cite{Xu2018b} layer, we use a $2$-layer MLP with ReLU activations, and batch normalization right after each GIN layer. 

\section{Further Results with SplineCNN} \label{sec:spline}

For the sake of completeness, We show additional results with the SplineCNN \cite{fey2018splinecnn} operator to validate the proposed block. We report the performance on the shape reconstruction benchmark. SplineCNN is conceptually similar by definition to MoNet \cite{Monti2017}, with a kernel function $g_{\mathbf{\Theta}}(\mathbf{u}_{i,j})$ represented on the tensor product of weighted B-Spline functions, that takes as input relative pseudo-coordinates $\mathbf{u}_{i,j}$. SplineCNN and MoNet both leverage the advantages of attention mechanisms to learn intrinsic features. To follow the definitions in Section 2 in the paper, we formulate the SplineCNN convolution as

\begin{equation}\label{eq:spline}
       \mathbf{x}^{(k)}_i = \frac{1}{|\mathcal{N}(i)|} \sum_{j \in
        \mathcal{N}(i)} \mathbf{x}^{(k-1)}_j \cdot
        g_{\mathbf{\Theta}}(\mathbf{u}_{i,j})    .
\end{equation}

\noindent Let $\mathbf{m} = (m_1, \ldots, m_d)$ the $d$-dimensional kernel size. For 3D data, the number of trainable weight matrices is $M=\prod_{i=1}^d m_i=m^3$, with equal kernel size in each three dimension. 

We show the results (Figure \ref{fig:spline}) obtained with SplineCNN and kernel sizes $m=1, \ldots, 5$. We fix the B-Spline degree to $1$, for both with and without affine skip connections\footnote{It should be noted that the implementation of SplineCNN provided by the author already uses a separate weight for the center vertex. To allow for a fair assessment of the affine skip connections, we tacitly assumed here that the propagation of SplineCNN is only based on Eq. \ref{eq:spline}}. The rest of the experimental setup and hyperparameters is identical to Section \ref{sec:setup}. Clearly, as shown in Figure \ref{fig:spline}, the performance of \textit{Aff}-SplineCNN is consistently better than that of SplineCNN, achieving the smallest error of all models at $0.241$ with kernel size 5 in \textit{each dimension} (\ie $125$ in total as the growth rate is cubical). Interestingly, SplineCNN (\textit{Aff}-SplineCNN) does not outperform MoNet (\textit{Aff}-MoNet) when the number of weight matrices is the same. For instance, for $M=8$, the mean Euclidean errors of MoNet and \textit{Aff}-MoNet are $0.531$ and $0.397$ respectively, whereas the mean Euclidean errors of SplineCNN and \textit{Aff}-SplineCNN are $0.605$ and $0.501$.






\end{document}